\documentclass[10pt,journal,compsoc]{IEEEtran}
\ifCLASSOPTIONcompsoc
  \usepackage[nocompress]{cite}
\else
  \usepackage{cite}
\fi

\usepackage{ifsym}
\usepackage{times}
\usepackage{epsfig}
\usepackage{graphicx}
\usepackage{lipsum}
\usepackage{amsmath}
\usepackage{multirow}
\usepackage{amssymb}
\usepackage[numbers,sort]{natbib} 
\usepackage{booktabs}
\usepackage{subcaption}
\usepackage{algorithm}
\usepackage{algorithmicx}
\usepackage{algpseudocode}
\usepackage{tikz}
\usepackage{diagbox} 

\usepackage[pagebackref=true,breaklinks=true,letterpaper=true,colorlinks,bookmarks=false]{hyperref}
\usetikzlibrary{positioning}

\def\spacetablecapt{\vspace{2pt}}
\def\spacesubsection{\vspace{0pt}}
\def\spacesection{\vspace{0pt}}
\def\newpara{\vspace{5pt}}

\begin{document}

\title{Deep Audio-Visual Speech Recognition}

\author{Triantafyllos Afouras, Joon Son Chung, Andrew Senior, Oriol Vinyals, Andrew Zisserman
\IEEEcompsocitemizethanks{
  \IEEEcompsocthanksitem{T.~Afouras and J.~S.~Chung are with the University of Oxford. \ \protect \\
  E-mail:\texttt{\{afourast,joon\}{@}robots.ox.ac.uk}}
  \IEEEcompsocthanksitem{A. Senior and O. Vinyals are with Google DeepMind. \ \protect \\
  E-mail:\texttt{\{vinyals,andrewsenior\}{@}google.com}}
  \IEEEcompsocthanksitem{A. Zisserman is with the University of Oxford and Google DeepMind. \ \protect \\
  E-mail:\texttt{az{@}robots.ox.ac.uk}}
     } 
\thanks{The first two authors contributed equally to this work.} 
}

%
%

\IEEEtitleabstractindextext{%
\begin{abstract}
   The goal of this work is to recognise phrases
   and sentences being spoken by a talking face, 
   with or without the audio. 
   Unlike previous works that have
   focussed on recognising a limited number of words
   or phrases, we tackle lip reading as an 
   {\em open-world} problem -- unconstrained natural language sentences, and in the wild videos.

   Our key contributions are: 
   (1) we compare two  models for lip reading, one using a CTC loss, and the other using a sequence-to-sequence loss.
Both models are built on top of the transformer self-attention architecture;
   (2) we investigate to what extent lip reading is complementary to audio speech recognition, especially when the audio
signal is noisy;
   (3) we introduce and publicly release a new 
    dataset for audio-visual speech recognition, LRS2-BBC,  consisting of
   thousands of natural sentences from British television.

The models that we train surpass the performance of
all previous work on a lip reading benchmark dataset by a
significant margin.  
\end{abstract}

}

\maketitle

\IEEEdisplaynontitleabstractindextext

%
\IEEEpeerreviewmaketitle

\IEEEraisesectionheading{\section{Introduction}\label{sec:introduction}}

%
%
%
%
\IEEEPARstart{L}{ip reading}, 
the ability to recognize what is being said from visual
information alone, is an impressive skill, and very challenging
for a novice. It is inherently ambiguous at the word level due to 
homophones --
different characters that produce exactly the same lip sequence ({\it e.g.}\
`p' and `b'). However, such ambiguities can be resolved to an
extent using the context of neighboring words in a sentence, and/or a
language model.

A machine that can lip read opens up a host of applications:
`dictating' instructions or messages to a phone in a noisy
environment; 
transcribing and re-dubbing archival silent films;  
resolving multi-talker simultaneous speech;
and, improving the 
performance of automated speech recognition in general.

That such automation is now possible is due to two developments that
are well known across computer vision tasks: the use of deep neural
network models~\cite{Krizhevsky12,Simonyan15,Szegedy15};
 and, the availability of a large scale dataset for
training~\cite{Russakovsky15}.  
In this case, the lip reading models are based on recent
encoder-decoder architectures that have been developed for speech recognition and
machine
translation~\cite{Graves06,Graves14,Bahdanau15,Chan15,Sutskever14}. 

The objective of this paper is to develop neural transcription architectures for lip reading sentences.
We compare two models: 
one using a {\em Connectionist Temporal Classification} (CTC) loss~\cite{Graves06}, and the other using a 
{\em sequence-to-sequence} (seq2seq) loss~\cite{Sutskever14,cho2014learning}.
Both models are based on
the transformer self-attention architecture~\cite{Vaswani2017}, so that the advantages and
disadvantages of the two losses can be compared head-to-head, with as much of the rest
of the architecture in common as possible.
The dataset developed in this paper to train and evaluate the models,
 are  based on thousands of hours of
videos that have talking faces together with
subtitles of what is being said.

We also investigate how lip reading can contribute to {\em audio}
based speech recognition. There is a large literature on this
contribution, particularly in noisy environments, as well as the
converse where some derived measure of audio can contribute to lip
reading for the deaf or hard of hearing. To investigate this aspect
we train a model to recognize characters from both audio and visual
input, and then systematically disturb the audio channel. 

Our models output at the character level.
In the case of the CTC, these outputs are independent of each other.
In the case of the sequence-to-sequence loss a language model is learnt implicitly, and
the architecture incorporates a novel dual attention mechanism that
can operate over visual input only, audio input only, or both. The
architectures are described in Section~\ref{sec:arc}.
Both models are decoded with a beam search, in which we can optionally incorporate an external
language model.

Section~\ref{sec:dataset}, we describe the generation and statistics
of a new large scale dataset, {\em LRS2-BBC}, 
that is used to train and evaluate
the models. The dataset contains talking faces together with subtitles of what is
said.  The videos contain faces `in the wild' with a significant
variety of pose, expressions, lighting, backgrounds and ethnic
origin.  Section~\ref{sec:training} describes the network training,
where we report a form of curriculum learning that is used to
accelerate training.  Finally, Section~\ref{sec:exp} evaluates the
performance of the models, including for visual (lips) input only, for
audio and visual inputs, and for synchronization errors between the
audio and visual streams.


\noindent\textbf{On the content:} This submission is 
based on the conference paper~\cite{Chung17}.
We replace the WLAS model in the original paper with
two variants of a Transformer-based model~\cite{Vaswani2017}. 
One variant was published in~\cite{Afouras18b}, and the second variant (using the CTC loss) is an original
contribution in this paper.
We also update the visual front-end
with a ResNet-based one proposed by~\cite{Stafylakis17}.
The new front-end and back-end architectures contribute to
over 22\% absolute improvement in Word Error Rate (WER) over the
model proposed in \cite{Chung17}.
Finally, we publicly release a new dataset, LRS2-BBC, that
supersedes the original LRS dataset in~\cite{Chung17}
which could not be made public due to license restrictions.

\begin{figure}[t] 
      \centering
              \includegraphics[width=\linewidth]{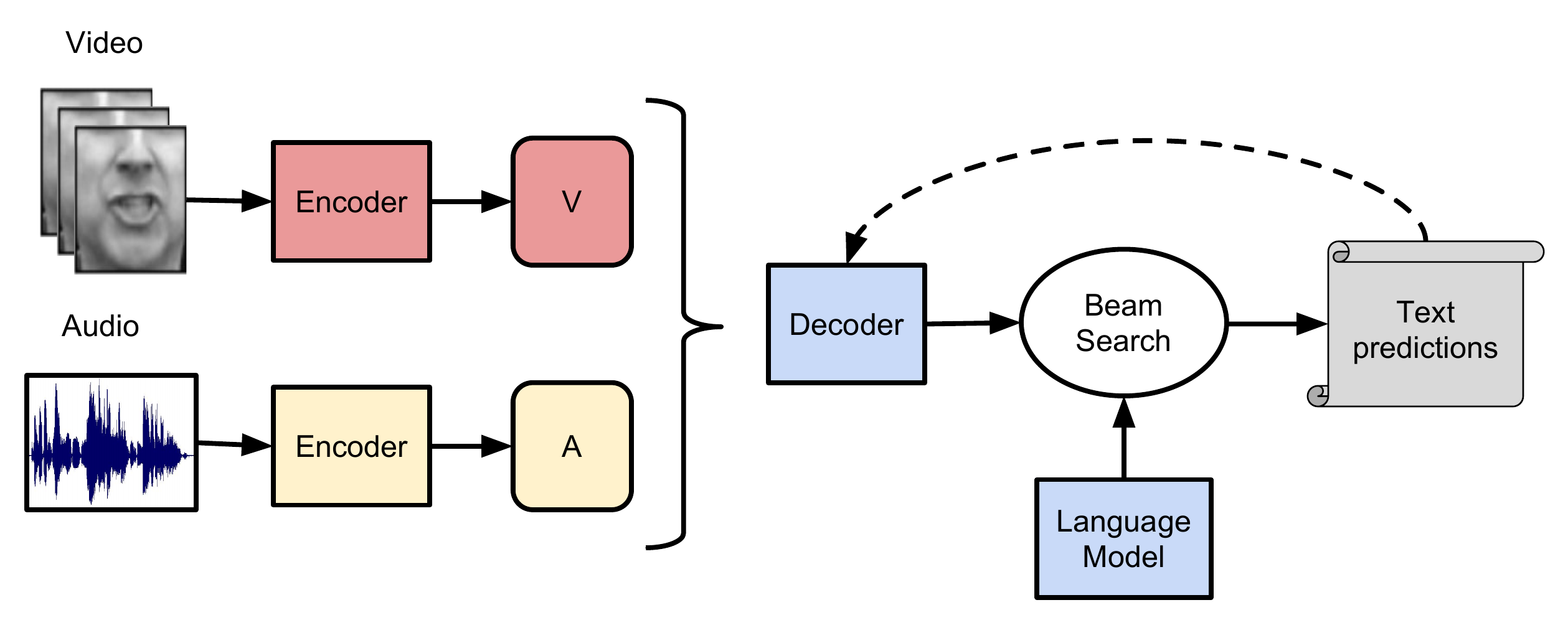}
              \caption{ Outline of the audio-visual speech recognition pipeline. }
              \label{fig:pipeline_small}
\end{figure}

\section{Background}
\spacesubsection

\subsection{CTC vs sequence-to-sequence architectures}

For the most part, end-to-end deep learning approaches for sequence prediction
 can be divided into two types.
 
The first type uses a neural network as an emission model which outputs the likelihood of each
output symbol ({\em e.g.}~phonemes) given the input sequence ({\em e.g.}~audio). These methods generally
employ a second phase of decoding using a Hidden Markov Model \cite{Hinton12b}. One such version of this variant is the Connectionist Temporal Classification (CTC)~\cite{Graves06}, where the model
predicts frame-wise labels and then looks for the optimal alignment
between the frame-wise predictions and the output sequence. 
The main weakness of CTC is that the output labels are not conditioned on
each other (it assumes each unit is independent), and hence a
language model is  employed as a post-processing step. Note that some
alternatives to jointly train the two step process has been proposed \cite{graves2012sequence}.
Another limitation of this approach is that it assumes a monotonic ordering between input and output sequences. This assumption is suitable for ASR 
and transcription for example, but not for 
machine translation.

The second type is sequence-to-sequence models~\cite{Sutskever14,cho2014learning} (seq2seq) that first read
all of the input sequence before predicting the output sentence.
A number of papers have adopted this approach
for speech recognition~\cite{Chorowski15,Chorowski14}: for example, 
Chan {\it et
al.}~\cite{Chan15}  proposes
an elegant sequence-to-sequence method
to transcribe audio signal to characters. 
Sequence-to-sequence decodes an  output symbol at time $t$ ({\em e.g.}\ character or word) conditioned on previous outputs $1,\ldots,t-1$. Thus,
unlike CTC-based models, the model implicitly learns a language model over output symbols, and no further processing is required.
However, it has been shown \cite{Chan15, Kannan17} that it is beneficial to incorporate
an external language model in the decoding of sequence-to-sequence models as
well. This way it is possible to leverage larger text-only corpora that
contain much richer natural language information than the limited aligned data used for training the
acoustic model.

Regarding architectures, while CTC-based or seq2seq approaches traditionally relied on recurrent networks,
recently there has been a shift towards purely convolutional models \cite{Bai18}. 
For example, fully convolutional networks have been used for ASR with CTC \cite{Wang17, Zhang} or a simplified
variant \cite{Collobert16, Liptchinsky17, Zeghidour17}.

\subsection{Related works}

\noindent{\bf Lip reading.}
There is a large body of work on lip reading using non deep learning
methods.  These methods are thoroughly reviewed in
\cite{Zhou14}, and we will not repeat this here.
A number of papers have used Convolutional Neural 
Networks (CNNs) to predict phonemes~\cite{Noda14}
 or visemes~\cite{Koller15}
from still images,
as opposed to recognising to full words or sentences. 
A {\it phoneme} is the smallest distinguishable unit
of sound that collectively make up a spoken word;
 a {\it viseme} is its visual equivalent. 

For recognising full words,
Petridis {\it et al.}~\cite{Petridis16} 
train an LSTM classifier on 
a discrete cosine transform (DCT)
and deep bottleneck features (DBF).
Similarly, Wand {\it et al.}~\cite{Wand16}
use an LSTM with HOG input features 
to recognise short phrases. 
The shortage of training data in lip reading 
presumably contributes to the continued use of
hand crafted features. 
Existing datasets consist of 
videos with only a small number of subjects,
and also a limited vocabulary ($<$60 words),
which is also an obstacle to progress.
Chung and Zisserman~\cite{Chung16}
tackles the small-lexicon problem by using faces in
television broadcasts to assemble the LRW dataset with
a vocabulary size of 500 words.
However, as with any word-level 
classification task,
the setting is still distant from the
real-world, given that the word
boundaries must be known beforehand.
Assael {\it et al.}~\cite{Assael16} 
uses a
CNN and LSTM-based network and (CTC)~\cite{Graves06} to compute the labelling.
This reports strong speaker-independent performance on the constrained
grammar and 51 word vocabulary of the GRID
dataset~\cite{Cooke06}.

A deeper architecture than LipNet \cite{Assael16} is used by~\cite{Stafylakis17}, who propose
a residual network with 3D convolutions to extract more  powerful
representations. The network is trained with a cross-entropy loss
to recognise words from the LRW dataset.
Here, the standard ResNet architecture \cite{He15} is modified to process 3D image
sequences by changing the first convolutional and pooling blocks from 2D to 3D.

In our earlier work~\cite{Chung17},  we proposed a WLAS sequence-to-sequence model 
based on the LAS ASR model of~\cite{Chan15} (the acronym WLAS are for Watch, Listen, Attend and Spell, 
and LAS for Listen, Attend and Spell). The WLAS model had a dual attention mechanism -- one for the visual (lip) stream,
and the other for the audio (speech) stream. It transcribed spoken 
sentences to characters, and could handle an input of
vision only, audio only,  or both. 

In independent and concurrent work, Shillingford {\it et al.}~\cite{Shillingford18}, design a lip
reading pipeline that uses a network which outputs phoneme probabilities and is trained with CTC
loss. At inference
time, they use a decoder based on finite state transducers to convert the phoneme distributions into word sequences. 
The network is trained on a very large scale lip
reading dataset constructed from YouTube videos and achieves a remarkable 40.9\% word error rate.

\noindent{\bf Audio-visual speech recognition.}
The problems of audio-visual speech recognition (AVSR)
and lip reading are
closely linked. 
Mroueh {\it et al.}~\cite{Mroueh15} employs
feed-forward Deep Neural Networks (DNNs) to perform
phoneme classification using a large
non-public audio-visual dataset.
The use of HMMs together with hand-crafted or pre-trained
visual features have proved popular
 --
 \cite{Tamura15} encodes
 input images using DBF; \cite{Galatas12} used DCT;
 and \cite{Noda15} uses a CNN pre-trained 
 to classify phonemes;
 all three combine these features with
 HMMs to classify spoken digits or isolated words.
 As with lip reading, there has been little
 attempt to develop AVSR systems that generalise
 to real-world settings.
 
Petridis {\it et al.}~\cite{Petridis18} use an extended version of the
architecture of~\cite{Stafylakis17} to learn representations from raw pixels and waveforms which
they then concatenate and feed to a bidirectional recurrent network that jointly models
the audio and video sequences and outputs word labels.



  \begin{figure*}[t] 
	\centering
		\includegraphics[width=\linewidth]{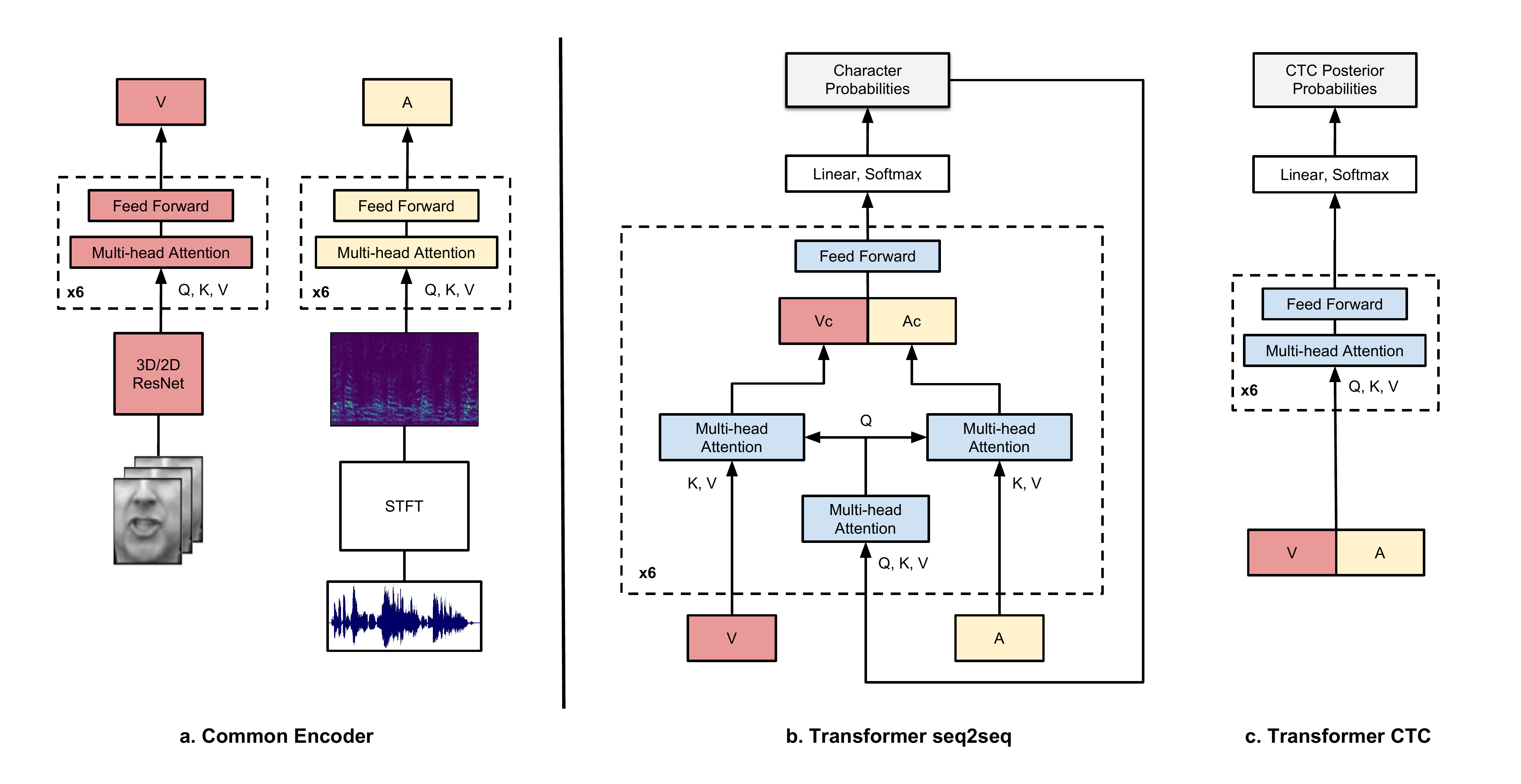}
                \caption{
                Audio-visual speech recognition models.
      \textbf{(a) Common encoder:}
                The visual image sequence is processed by a spatio-temporal ResNet, while the audio
                features are the spectrograms obtained by applying Short Time Fourier Transform
                (STFT) to the audio signal. Each modality is then encoded by a separate Transformer
                encoder. 
                \textbf{(b) TM-seq2seq}: a Transformer model. On every decoder layer, the video (V) and audio (A) encodings are attended to
       separately by independent multi-head attention modules. The context vectors produced for the
       two modalities, $V_c$ and $A_c$ respectively, are concatenated channel-wise and fed to the
       feed forward layers.  K, V and Q denote the Key, Value and Query tensors for the multi-head
       attention blocks. For the self-attention layers it is always $Q=K=V$, while for the
       encoder-decoder attentions, $K$ = $V$ are the encodings (V or A), while $Q$
       is the previous layer's output (or, for the first layer, the prediction of the
       network at the previous decoding step).
       \textbf{(c) TM-CTC}: Transformer CTC, a model composed of stacks of self-attention and
       feed forward layers, producing CTC posterior probabilities for every input frame. 
       For full details on the multi-head attention and feed forward blocks refer to Appendix \ref{ap:transformer_details}. 
    }
      \label{fig:architecture}
\end{figure*}

\section{Architectures}
\label{sec:arc}
\spacesection

In this section, we describe model architectures for audio-visual speech recognition, for which we explore two variants, based on the
recently proposed Transformer model~\cite{Vaswani2017}: i) an encoder-decoder
attention structure for training in a seq2seq manner and ii) a stack of self-attention blocks for training with CTC loss.
The architecture is outlined in Figure~\ref{fig:architecture}.
The general model receives two input streams, one for video (V) and one for audio (A).  

\subsection{Audio Features} 
For the acoustic representation we use 321-dimensional spectral magnitudes, computed with a 40ms
window and 10ms hop-length, at a 16 kHz sample rate.
Since the video is sampled at $25$ fps ($40$ ms per frame), every
video input frame corresponds to $4$ acoustic feature frames. We concatenate the audio
features in groups of $4$, in order to reduce the input sequence length as is common for stable CTC training~\cite{sak15, Chiu17},
while at the same time achieving a common temporal-scale for both modalities.

\subsection{Vision Module (VM)} 
\begin{figure*}[th!]
\centering

\includegraphics[width=.246\textwidth]{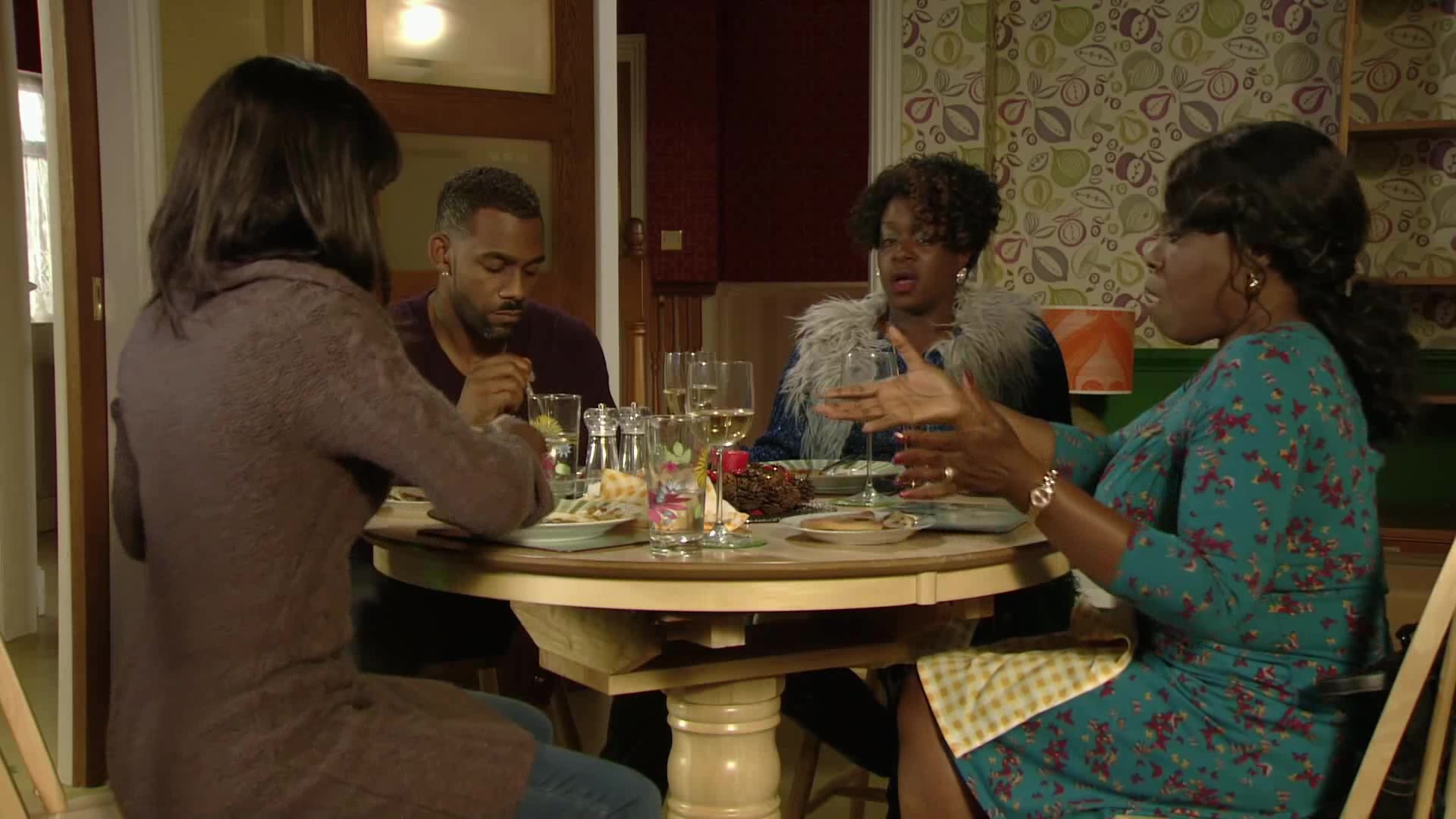} 
\includegraphics[width=.246\textwidth]{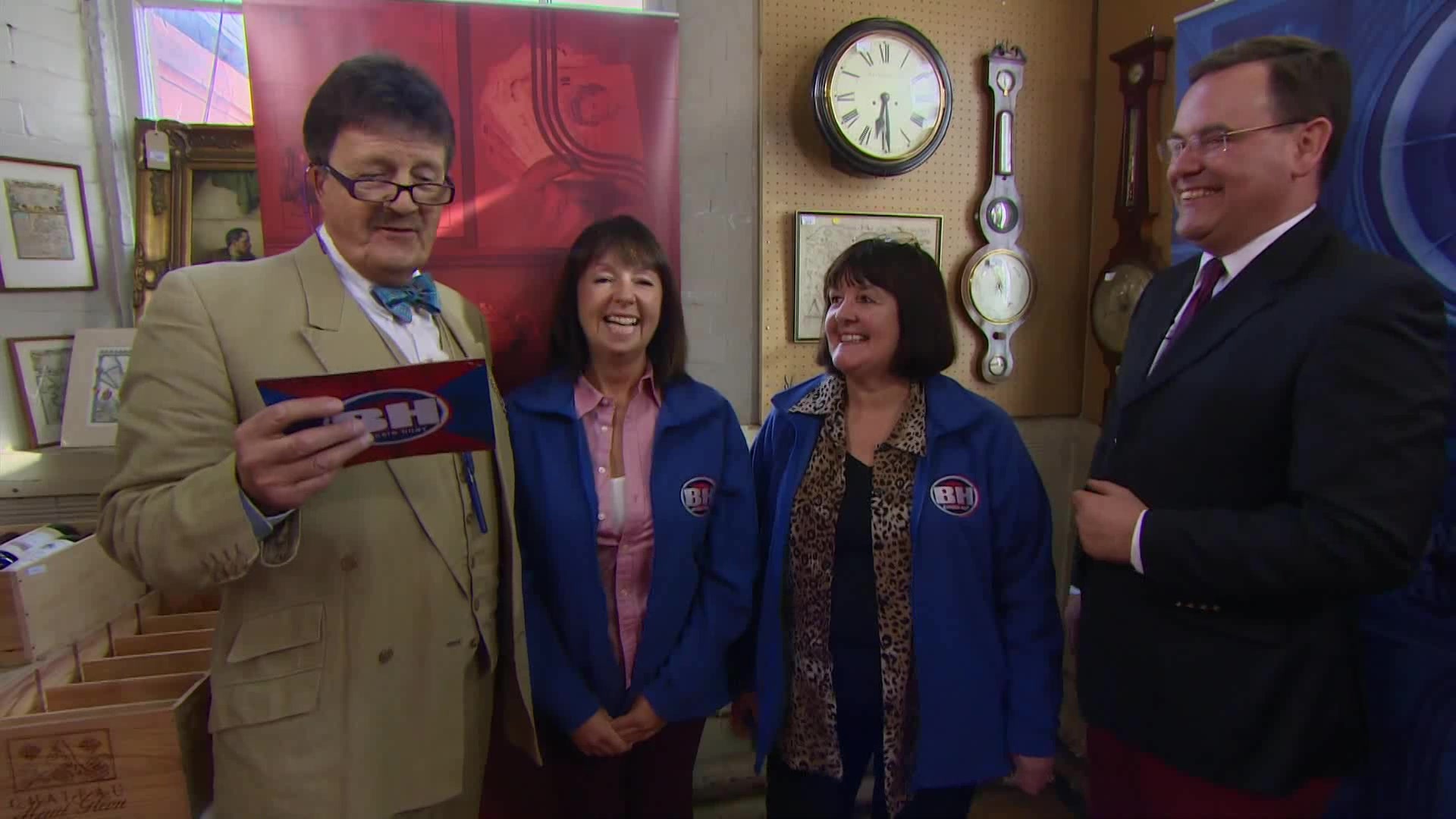}
\includegraphics[width=.246\textwidth]{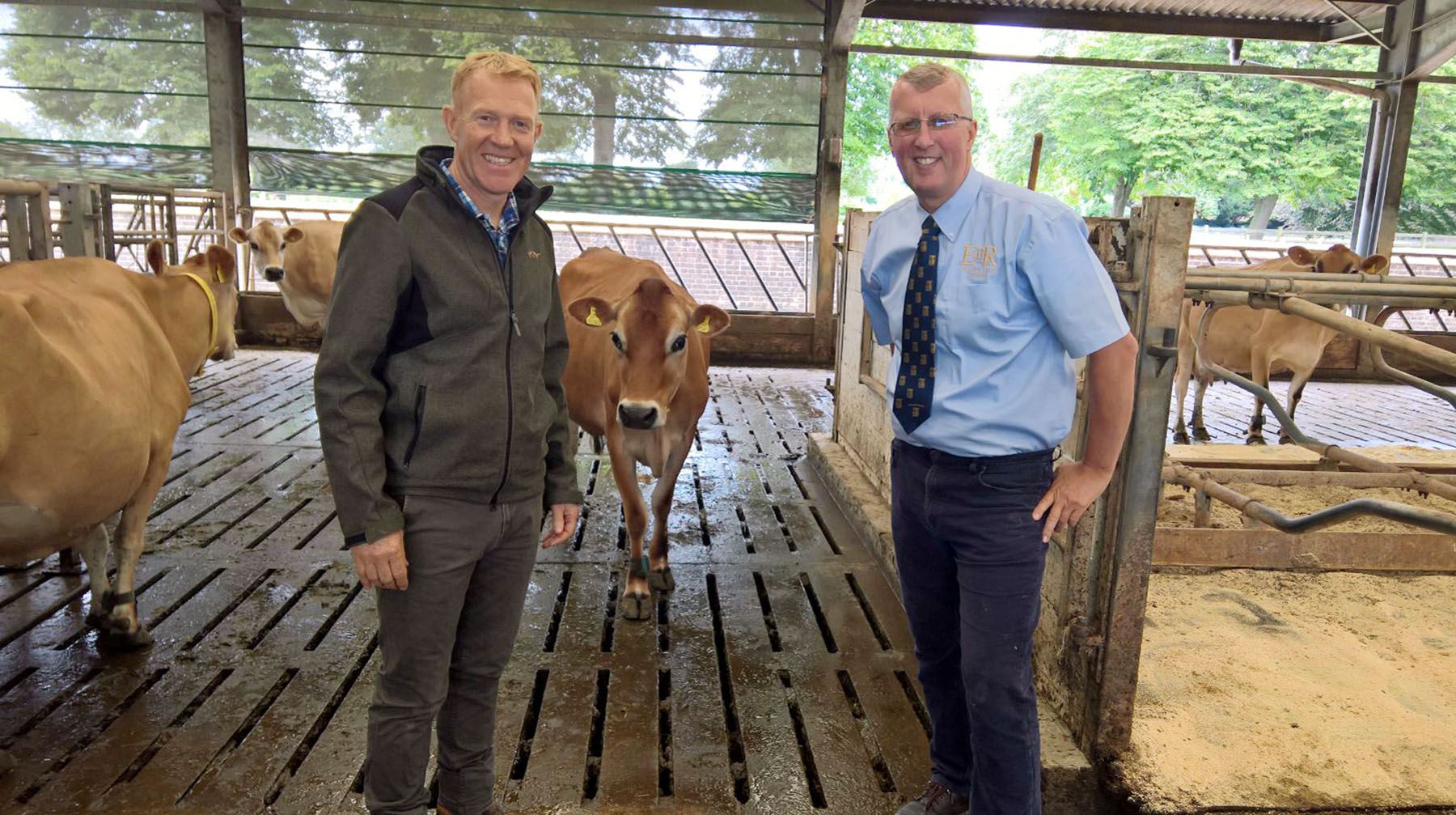}
\includegraphics[width=.246\textwidth]{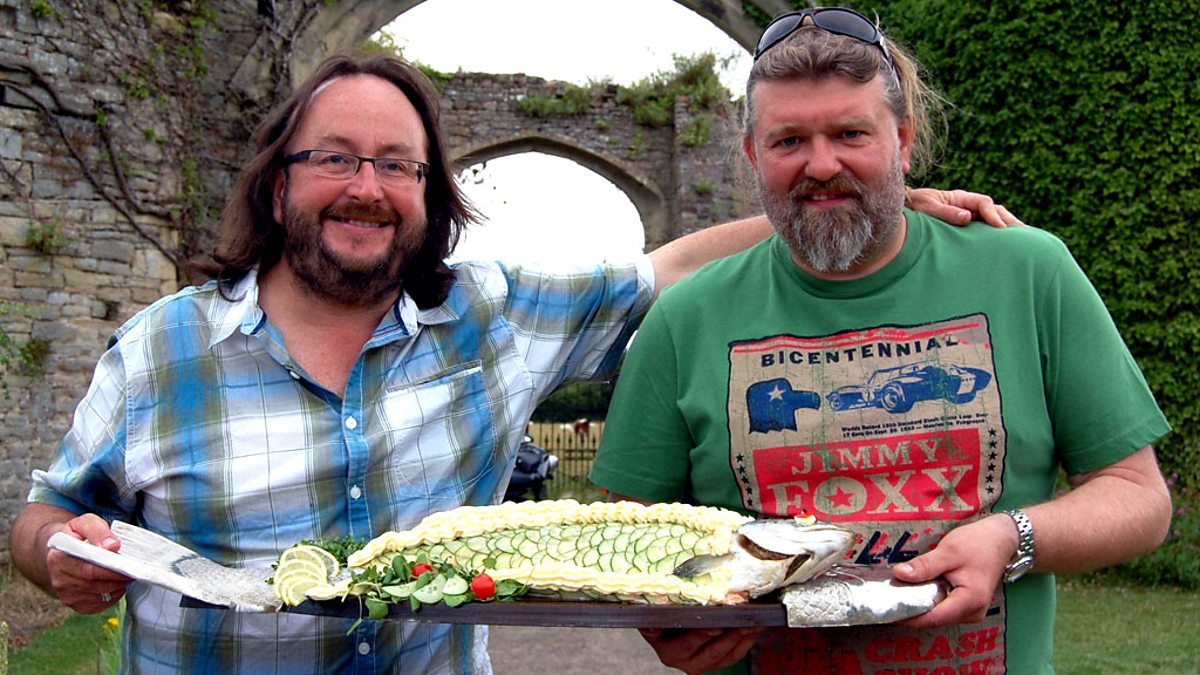} 

\vspace{5pt}

\includegraphics[width=0.058\textwidth]{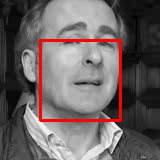}
\includegraphics[width=0.058\textwidth]{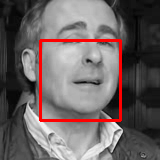}
\includegraphics[width=0.058\textwidth]{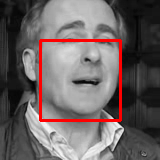}
\includegraphics[width=0.058\textwidth]{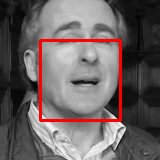}
\includegraphics[width=0.058\textwidth]{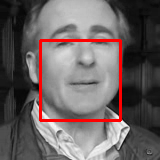}
\includegraphics[width=0.058\textwidth]{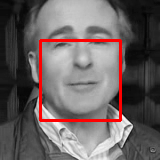}
\includegraphics[width=0.058\textwidth]{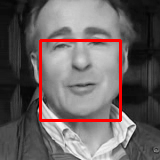}
\includegraphics[width=0.058\textwidth]{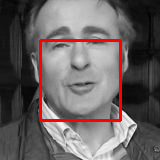} 
\includegraphics[width=0.058\textwidth]{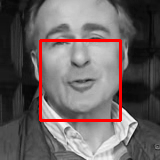}
\includegraphics[width=0.058\textwidth]{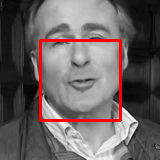}
\includegraphics[width=0.058\textwidth]{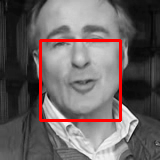}
\includegraphics[width=0.058\textwidth]{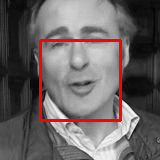}
\includegraphics[width=0.058\textwidth]{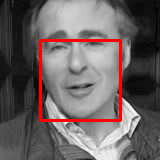}
\includegraphics[width=0.058\textwidth]{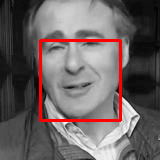}
\includegraphics[width=0.058\textwidth]{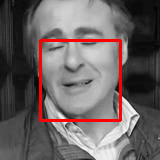}
\includegraphics[width=0.058\textwidth]{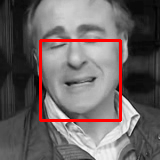}

\vspace{5pt}

\includegraphics[width=0.058\textwidth]{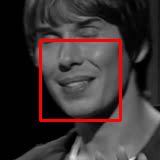}
\includegraphics[width=0.058\textwidth]{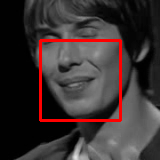}
\includegraphics[width=0.058\textwidth]{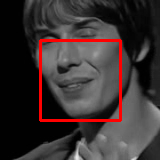}
\includegraphics[width=0.058\textwidth]{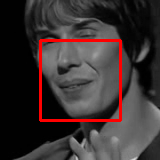}
\includegraphics[width=0.058\textwidth]{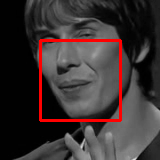}
\includegraphics[width=0.058\textwidth]{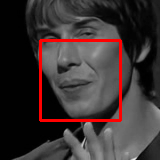}
\includegraphics[width=0.058\textwidth]{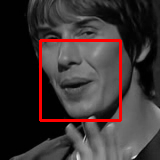}
\includegraphics[width=0.058\textwidth]{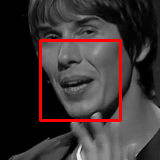} 
\includegraphics[width=0.058\textwidth]{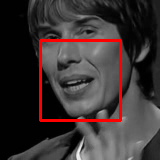}
\includegraphics[width=0.058\textwidth]{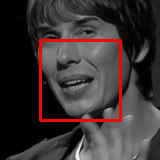}
\includegraphics[width=0.058\textwidth]{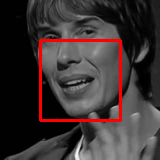}
\includegraphics[width=0.058\textwidth]{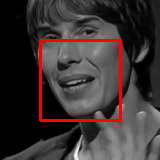}
\includegraphics[width=0.058\textwidth]{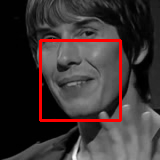}
\includegraphics[width=0.058\textwidth]{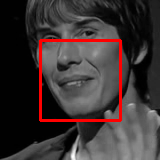}
\includegraphics[width=0.058\textwidth]{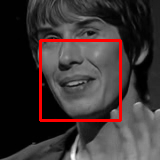}
\includegraphics[width=0.058\textwidth]{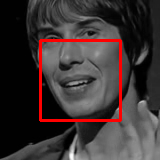}

\caption{{\bf Top:} Original still images from videos used in the making of the LRS2-BBC dataset.
 {\bf Bottom:} The mouth motions from two different speakers.
The network sees the areas inside the red squares.}
\label{fig:pm} 
\end{figure*}

The input images are 224$\times$224 pixels, sampled at 25 fps and contain the speaker's face. 
We crop a 112$\times$112 patch covering the region around the mouth, as shown in
Figure~\ref{fig:pm}.
To extract visual features representing the lip movement, we use a spatio-temporal visual front-end 
that is based on \cite{Stafylakis17}. 
The network applies 3D convolutions on the input image sequence, with a filter width of $5$ frames, 
followed by a 2D ResNet that gradually decreases the spatial dimensions with depth. The layers are
listed in full detail in Appendix \ref{ap:visual_frontend}.
For an input sequence of $T \times H \times W$ frames, the output is a $T \times \frac{H}{32} \times
\frac{W}{32} \times 512$ tensor ({\em i.e.}\ the temporal resolution
is preserved) that is then average-pooled over the spatial dimensions, yielding a $512$-dimensional feature
  vector for every input video frame.  

  \subsection{Common self-attention Encoder}
  Both variants that we consider use the same self-attention-based encoder architecture.
  The encoder is a stack of multi-head self-attention layers, where the input tensor serves as the
  query, key and value for the attention at the same time.
A separate encoder is used for each modality as shown in Figure~\ref{fig:architecture}~(a).
  The information about the sequence order of the 
  inputs is fed to the model via fixed positional embeddings in the form of sinusoid functions.

  \subsection{Sequence-to-sequence Transformer (TM-seq2seq)}
  In this variant, separate attention heads are used for attending on the video and audio embeddings.
  In every decoder layer, the resulting video and audio contexts are concatenated over the channel dimension and propagated to
  the feedforward block.
  The  attention mechanisms for both modalities receive as queries the output of the previous
  decoding layer (or the decoder input in the case of the first layer).
  The decoder produces character probabilities which are directly matched to the ground truth labels
  and trained with a cross-entropy loss. 
  More details about the multi-head attention and
  feed-forward building blocks are given in Appendix \ref{ap:transformer_details}.

  \subsection{CTC Transformer (TM-CTC)}
  The TM-CTC model concatenates the video and audio encodings and propagates the result
  through a stack of self-attention / feedforward blocks, same as the one used in the encoders. The outputs of the network are the CTC
  posterior probabilities for every input frame and the whole stack is trained with CTC loss.

  \subsection{External Language Model (LM) }
  For decoding both variants, during inference, we use a character-level language model.
   It is a recurrent network with 4 unidirectional layers of 1024 LSTM cells each.
   The language model is trained to predict one character at a time, receiving only the previous
   character as input.
   Decoding for both models is performed with a left-to-right beam search where the LM log-probabilities are combined 
   with the model's outputs via shallow fusion \cite{Kannan17}. 
   More details on decoding are given in Appendices \ref{ap:seq2seq_dec} and \ref{ap:ctc_dec}.

  \subsection{Single modality models }
  The audio-visual models described in this section can be used when only one of the two modalities is
  present. Instead of concatenating the attention vectors for TM-seq2seq or the encodings for TM-CTC, only the
  vector from the available modality is used.


\section{Dataset}
\label{sec:dataset}
\spacesection

In this section, we describe the multi-stage pipeline for 
automatically generating a large-scale dataset, {\em LRS2-BBC},  for
audio-visual speech recognition. 
Using this pipeline, we have been
able to collect thousands of hours of spoken sentences and
phrases along with the corresponding facetrack.
We use a variety of BBC programs from Dragon's Den to Top Gear and Countryfile.

\begin{figure}[ht]
\centering 
\includegraphics[width=1\linewidth]{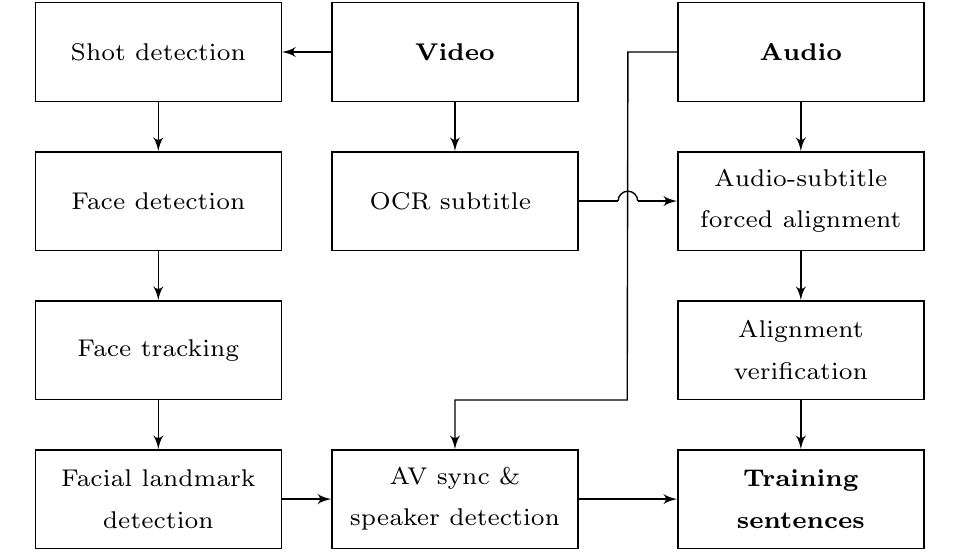}
\caption{Pipeline to generate the dataset.}
\label{fig:pipeline} 
\end{figure}

The processing pipeline is summarised in Figure~\ref{fig:pipeline}. 
Most of the steps are based on the methods
described in \cite{Chung16} and \cite{Chung16a},
but we give a brief sketch of the method here.

\newpara\noindent\textbf{Video preparation.} 
A CNN face detector based on the
Single Shot MultiBox Detector (SSD)~\cite{Liu16}
is used to detect face appearances in the individual frames.
Unlike the HOG-based detector~\cite{King09} used
by previous works, the SSD detects faces 
from all angles, and shows a more robust performance whilst
being faster to run.

The shot boundaries are determined by comparing color histograms
across consecutive frames~\cite{Lienhart01}.
Within each shot, face tracks are generated from face detections
based on their positions, as feature-based
trackers such as KLT~\cite{Lucas81} often fail
when there are extreme changes in viewpoints.

\newpara\noindent\textbf{Audio and text preparation.}
The subtitles in television
are not broadcast
in sync with the audio.
The Penn Phonetics Lab Forced
Aligner~\cite{Yuan08} 
is used to
force-align the subtitle to the audio signal.
Errors exist in the alignment as the transcript is
not verbatim --
therefore the aligned labels are filtered by
checking against the commercial IBM Watson Speech to Text
service.

\newpara\noindent\textbf{AV sync and speaker detection.}
In broadcast videos, the audio and the video streams can be out of
sync by up to around one second, which can cause problems when
the facetrack corresponding to a sentence is being extracted.
A multi-view adaptation~\cite{Chung17a} of 
the two-stream network described in \cite{Chung16a} is used to 
synchronise the two streams. 
The same network is also used to determine 
which face's lip movements match the audio,
and if none matches, the clip is rejected
as being a voice-over.

\newpara\noindent\textbf{Sentence extraction.} 
The videos are divided
into individual sentences/ phrases using the punctuations
in the transcript. 
The sentences are separated by full stops, commas 
and question marks; and
are clipped to 100 characters or 10 
seconds, due to GPU memory constraints.
We do not impose any restrictions on the vocabulary size.

The LRS2-BBC dataset is divided into development (train/val) and test
sets according to broadcast date.  The dataset also has a ``{\em
pre-train}'' set that contains sentence excerpts which may be shorter
or longer than the full sentences included in the development set,  and
are annotated with the alignment boundaries of every word.  The
statistics of these sets are given in Table~\ref{table:datastat}.
The table also compares the {\it `Lip Reading Sentences'} (LRS) series of datasets 
to the largest existing public datasets.
In addition to LRS2-BBC, we use MV-LRS and LRS3-TED for training and evaluation.

\begin{table*}[th!]
\setlength{\tabcolsep}{9pt}
\centering
\footnotesize
\begin{tabular}{ l  c  l  c  r   r  r  r r }
  \toprule
  \textbf{Dataset} & \textbf{Source} & \textbf{Split} & \textbf{Dates} & \textbf{\# Spk.} &
  \textbf{\# Utt.} & \textbf{Word inst.} & \textbf{Vocab} & \textbf{\# hours}  \\ 
  \midrule 
  
   GRID~\cite{Cooke06} & - & - & - 					& 51 & 33,000 & 165k & 51 & 27.5\\ \midrule
   MODALITY~\cite{Czyzewski17} & - & - & - 		& 35 &  5,880 & 8,085 & 182 & 31 \\ 
   \midrule
   
  \multirow{2}{*}{LRW~\cite{Chung16}}  &  \multirow{2}{*}{BBC}  & Train-val & 01/2010 - 12/2015 & -
  & 514k & 514k & 500 & 165 \\ 
   &  & Test & 01/2016 - 09/2016 & - & 25k & 25k & 500  & 8\\
   \midrule 

  \multirow{2}{*}{LRS~\cite{Chung17}} $\dag$ &  \multirow{2}{*}{BBC}  & Train-val & 01/2010 -
  02/2016 & - & 106k & 705k & 17k & 68\\ 
  &  & Test & 03/2016 - 09/2016 & - & 12k & 77k & 6,882 & 7.5\\
   \midrule 

 \multirow{3}{*}{MV-LRS~\cite{Chung17a}} $\dag$ & \multirow{3}{*}{BBC} 
  &  Pre-train   &  01/2010 - 12/2015 &  - & 430k   & 5M     & 30k   & 730 \\ 
  &  & Train-val  & 01/2010 - 12/2015   & - &  70k & 470k   & 15k  & 44.4  \\ 
  &  & Test       & 01/2016 - 09/2016   &  - & 4,305 &  30k & 4,311 & 2.8 \\ 
  \midrule

 \multirow{4}{*}{{\bf LRS2-BBC}} & \multirow{4}{*}{BBC} 
 &   Pre-train  & 01/2010 - 02/2016   &  - & 96k & 2M & 41k & 195 \\ 
  &  &  Train-val  & 01/2010 - 02/2016    &  - & 47k & 337k & 18k & 29 \\ 
  &  & Test        & 03/2016 - 09/2016   &  - & 1,243 &  6,663 & 1,693 & 0.5 \\ 
  &  & Text-only        & 01/2016 - 02/2016 &  -  &  8M & 26M   & 60k & - \\ 
  \midrule

 \multirow{4}{*}{{LRS3-TED~\cite{Afouras18d}}} &   \multirow{4}{*}{\shortstack{TED \& \\ TEDx\\(YouTube)}} 
 &   Pre-train  & -  & 5,075 & 132k & 4.2M & 52k & 444 \\ 
  & &  Train-val                                       & -   & 3,752 & 32k & 358k & 17k & 30  \\ 
  & & Test                                               & - & 452 & 1,452 &  11k & 2,136 & 1  \\ 
  &  & Text-only        &  - &  5,075 & 1.2M  &  7.2M & 57k & - \\ 
  \midrule 

\end{tabular} 
\normalsize
\spacetablecapt
\caption{Statistics on the {\bf Lip Reading Sentences (LRS) audio-visual datasets,} and other existing large-scale lip reading datasets. Division of training, validation and test data; and the number of utterances, number of word instances and vocabulary size of each partition.
{\bf Utt:} Utterances.
$\dag$: Not available to the public due to license restrictions.
}
\label{table:datastat}
\end{table*}

\newpara\noindent\textbf{Datasets for training external language models}.
To train the language models used for evaluation on each audio-visual dataset, we use a text corpus containing the full subtitles
of the videos from which the dataset's training set was generated. The text-only corpus contains $26M$ words.


\section{Training strategy}
\label{sec:training}
\spacesection

In this section, we describe the strategy used to effectively train the
models, making best
use of the limited amount of data available.
The training proceeds in four stages: i) the visual front-end module is trained;
ii) visual features are generated for all the training data using the vision module; 
iii) the sequence processing module is trained on the frozen visual features; iv) the whole network is
trained end-to-end.

\subsection{Pre-training visual features}
\label{subsec:training}

We pre-train the visual front-end on word excerpts from the
MV-LRS~\cite{Chung17a} dataset,
using a 2-layer temporal convolution back-end to classify every
clip with a word label similarly to \cite{Stafylakis17}.
We perform data augmentation in the form of horizontal flipping, 
removal of random frames~\cite{Assael16,Stafylakis17}, 
and random shifts of up to $\pm5$ pixels in the spatial dimensions and 
of $\pm2$ frames in the temporal dimension.

\subsection{Curriculum learning} \label{sec:curriculum} 
\spacesubsection

Sequence to sequence learning has been reported to converge very slowly when the number
of timesteps is large, because the decoder initially
has a hard time extracting the relevant information
from all the input steps \cite{Chan15}.
Even though our models do not contain any recurrent modules, 
we found it beneficial to follow a curriculum instead of immediately training on full sentences. 

We introduce a new strategy where we start training only on single
word examples, and then let the sequence length grow as the 
network trains. 
These short sequences are parts of the longer sentences in the dataset.
We observe that the rate of convergence on the training set 
is several times faster, 
while the curriculum also significantly reduces overfitting, presumably because it 
works as a natural way of augmenting the data.

The networks are first trained on the frozen features of the {\em pre-train} sets from MV-LRS, LRS2-BBC and LRS3-TED.
We deal with the difference in utterance lengths by zero-padding the sequences to a maximum length,
which we gradually increase.
We then separately fine-tune end-to-end on the {\em train-val} set of LRS2-BBC or
LRS3-TED, according to which set we are evaluating on.

\subsection{Training with noisy audio \& multi-modal training}
\label{sec:trainwithnoise}
\spacesubsection
The audio-only models are initially trained with clean input audio.
Networks with multi-modal inputs can often be dominated by one of the modes \cite{Feichtenhofer16}.
In our case we observe that for the audio-visual models the audio signal dominates, because 
speech recognition is a significantly
easier problem than lip reading. 
To help prevent this from happening,
we add babble noise with 0dB SNR to the audio stream with
probability $p_{n}=0.25$ during training.  

To assess and improve tolerance to audio noise, we then fine-tune the audio-only and audio-visual
models in a setting where babble noise with 0dB SNR is always added to the original audio. 
We synthesize the babble noise samples by mixing the signals of 20 different audio
samples from the LRS2-BBC dataset. 

\subsection{Implementation details}
\label{sec:implem}
\spacesubsection

The output size of the network is 40, accounting for
the 26 characters in the alphabet, the 10 digits, and tokens for \texttt{[space]} and \texttt{[pad]}.
For TM-seq2seq we use an extra \texttt{[sos]} token and for TM-CTC the \texttt{[blank]} token. We
do not model punctuation, as the transcriptions of the datasets do not contain any.

The TM-seq2seq is trained using teacher forcing -- we supply the ground truth
of the previous decoding step as the input to the decoder, while
during inference we feed back the decoder prediction. 

Our implementation is based on the TensorFlow library~\cite{Abadi16}
and trained on a single GeForce GTX 1080 Ti GPU with 11GB memory. 
The network is trained 
using the ADAM optimiser~\cite{kingma2014adam} with the default
parameters and an initial learning rate of $10^{-4}$, which is reduced by a factor of 2 every time the validation error
plateaus, down to a final learning rate of $10^{-6}$.
For all the models we use dropout with $p=0.1$ and label smoothing.



\section{Experiments}
\label{sec:exp}
\spacesection

In this section we evaluate and compare the proposed architectures
and training strategies.  We also compare our methods
to the previous state of the art.

We train as described in section \ref{sec:curriculum} and evaluate the fine-tuned models for
LRS2-BBC and LRS3-TED on the independent test set of the respective dataset.
The inference and evaluation procedures are described below.

\newpara\noindent{\bf Test time augmentation. } 
During inference we perform 9 random transforms (horizontal flipping of the video frames and spatial shifts up to $\pm5$ pixels)
on every video sample, and pass the perturbed sequences through the network, in addition to the original.
For TM-seq2seq we average the resulting logits whereas for TM-CTC we average the visual features.

\newpara\noindent{\bf Beam search. } 
Decoding is performed with beam search of width 35 for TM-Seq2seq and 100 for TM-CTC (the values
were determined on a held-out validation set from the {\em{train-val}} split of LRS2-BBC).

\begin{table}[t!]
\centering
\footnotesize
\begin{tabular}{ l c   r  r   r  r  }
 \toprule

 \multirow{2}{*}{\backslashbox{\textbf{Method\kern-1em}}{\textbf{Dataset \kern-1em}}}
   & &    \multicolumn{2}{c}{{\bf LRS2-BBC}} & \multicolumn{2}{c}{{\bf LRS3-TED}} \\
   \cmidrule(l r{1\tabcolsep}){3-4}              \cmidrule(l{1\tabcolsep} r){5-6}  
   & \textbf{M} & \textbf{} & \textbf{+ extLM} & \textbf{} & \textbf{+ extLM}  \\ 


 \midrule


    Google S2T$\dag$   & A & \multicolumn{2}{c}{20.9\%} & \multicolumn{2}{c}{10.4\%} \\
    WAS~\cite{Chung17}   & V & 70.4\% & - & - & - \\

    \addlinespace[0.7em]

    TM-CTC      & V    & 65.0\% & 54.7\%            & 74.7\% & 66.3\%\\ 
    TM-CTC      & A    & 15.3\% & 10.1\%            & 13.8\% & 8.9\%\\ 
    TM-CTC      & AV   & 13.7\%  & 8.2\%            & 12.3\% & 7.5\%\\ 

    \addlinespace[0.7em]

    TM-seq2seq & V  & 49.8\%    &48.3\%    & 59.9\%   & 58.9\%\\
    TM-seq2seq & A  & 10.5\%    & 9.7\%               & 9.0\%  & 8.3\%\\ 
    TM-seq2seq & AV &  9.4\%    & 8.5\%               & 8.0\%  & 7.2\%\\ 

 \midrule
 \addlinespace[0.5em]
    \multicolumn{6}{c}{{\bf Noisy}} \\ 
 \addlinespace[0.3em]
 
 Google S2T$\dag$   & A & \multicolumn{2}{c}{86.3\%} & \multicolumn{2}{c}{70.3\%} \\
    \addlinespace[0.7em]
    
    TM-CTC   & A  & 64.7\% & 53.4\%               & 65.6\% & 56.3\%\\ 
    TM-CTC   & AV & 33.5\% & 23.6\%               & 37.2\% & 27.7\%\\ 

    \addlinespace[0.7em]
    TM-seq2seq   & A  & 58.0\% & 57.4\%           & 60.5\% & 57.9\%\\ 
    TM-seq2seq   & AV & 35.9\% & 34.2\%           & 44.3\% & 42.5\%\\ 
 \bottomrule

\end{tabular} 
\normalsize
\spacetablecapt
\caption{
Word error rates (WER) on the LRS2-BBC and LRS3-TED datasets. 
The second column (M) specifies the input modalities: V, A, and AV denote
video-only, audio-only, and audio-visual models respectively, while
+ extLM denotes decoding with the external language model.
$\dag$ {\scriptsize \url{https://cloud.google.com/speech-to-text}}, accessed 3 July 2018.
 }
\label{table:mainresults}
\end{table}

\newpara\noindent{\bf Evaluation protocol. } 
For all experiments,
we report the Word Error Rate (WER) which is defined as $\mathtt{WER} = (S+D+I)/N$, 
where $S$, $D$ and $I$ are the number of substitutions, deletions,
and insertions respectively to get from the reference to the hypothesis,
and $N$ is the number of words in the reference.

\newpara\noindent{\bf Experimental setup. } 
The rest of this section is structured as follows: First we present
results on lip reading, where only the video is used as input.
We then use the full models for audio-visual speech recognition, where the video and audio are
assumed to be properly synchronised. To assess the robustness of our models in noisy environments we also train and
test in a setting where babble noise is artificially added to the utterances. 
Finally we present some experiments on non-synchronised video and audio. 
The results for all experiments are summarized in Table~\ref{table:mainresults}, where we report
word error rates depending on whether a language model is used during decoding or not.

\subsection{Lips only}
\label{sec:lips}
\spacesubsection

\newpara\noindent\textbf{Results.}
The best performing network is TM-seq2seq, which achieves a WER of $48.3\%$ on LRS2-BBC
when decoded with a language model, an absolute improvement of over $22\%$ compared to the previous 
 $70.4\%$ state-of-the-art~\cite{Chung17}. This model also sets a baseline for LRS3-TED at 58.9\%.

\begin{figure}[t] 
      \centering
              \includegraphics[width=\linewidth]{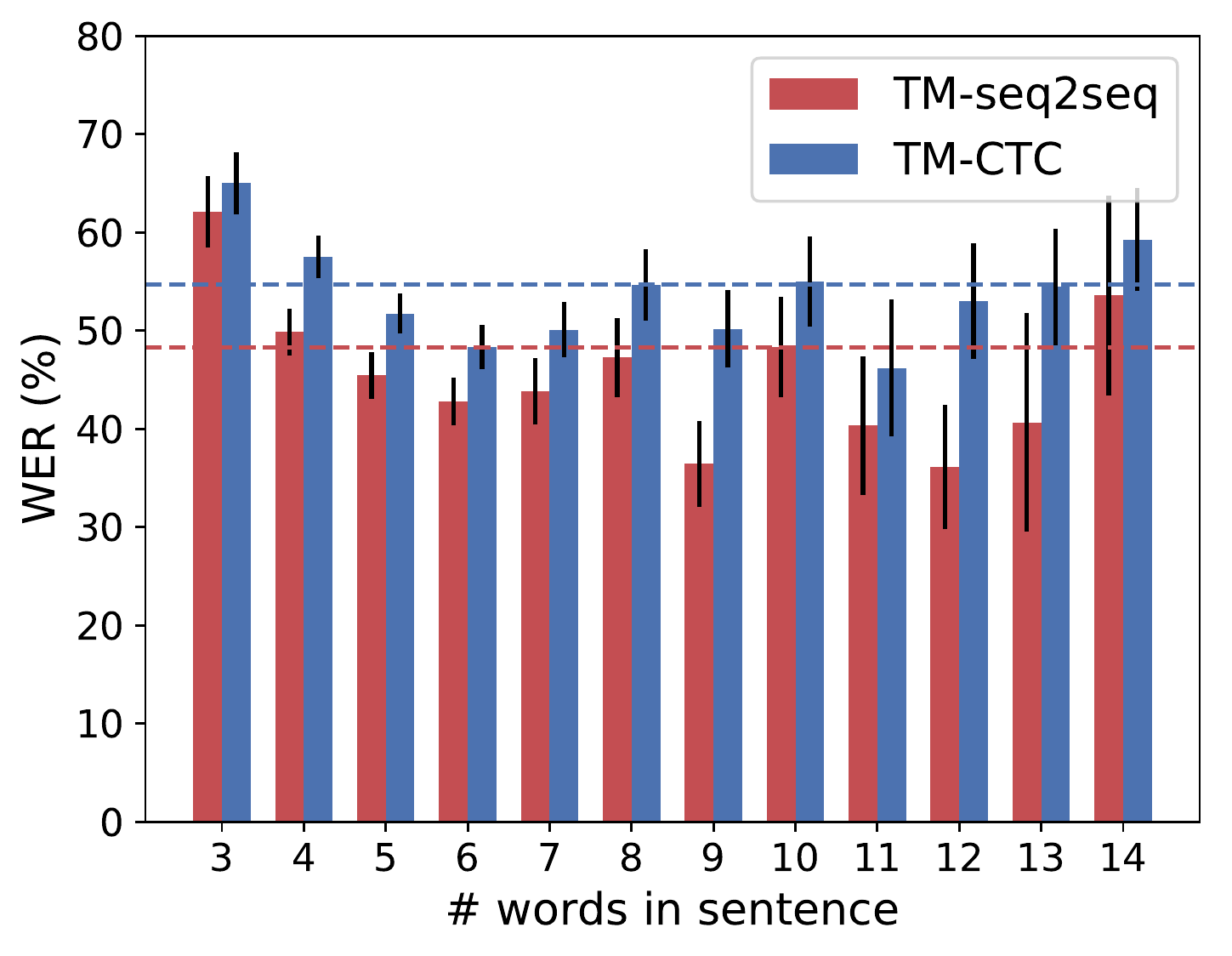}
              \caption{ Word error rate per number of words in the sentence for the video-only
                models, evaluated on the test set of LRS2-BBC. We exclude sentence sizes
                represented by less than 5 samples in the set (i.e. 15, 16 and 19 words). The dashed lines show the average WER over all the sentences.
                For both models, the WER is relatively uniform for different sentence sizes.
                However samples with very few words (3) appear to be more difficult, presumably
              because they provide less context. }
              \label{fig:wer_per_len}
\end{figure}

\begin{figure}[t]
\centering 

\begin{subfigure}[b]{\columnwidth}
  \includegraphics[width=1\linewidth]{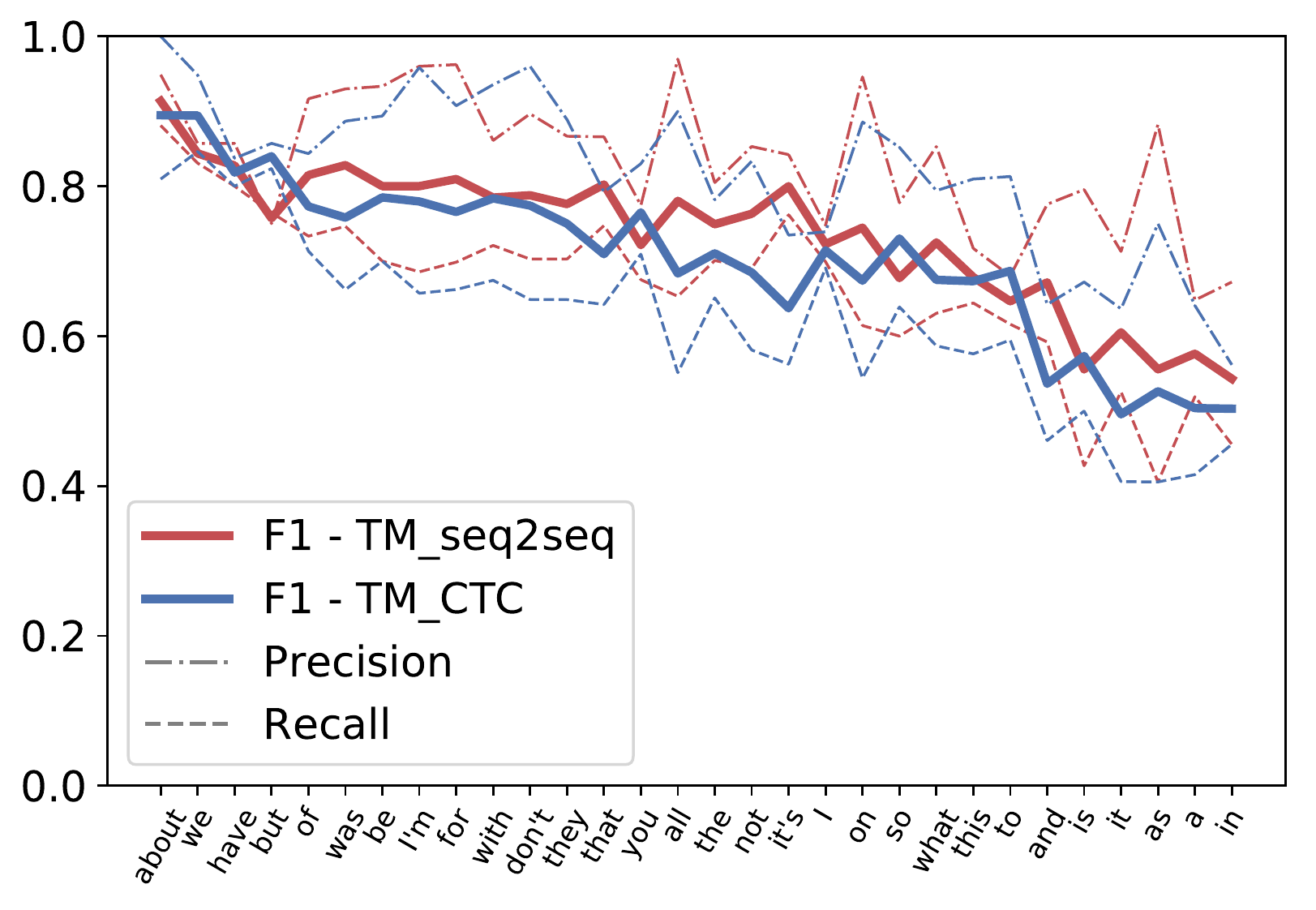}
 \end{subfigure}



 \caption{ Per word F1, Precision and Recall rates, on the 30 most common words in the LRS2-BBC
   test set, for the video-only models.
   The measures are calculated via the minimum edit-distance operations (details in Appendix \ref{ap:prec_recall}).
   For all words and both models, precision is higher than recall.
 }
\label{fig:f1_prec_rec} 
\end{figure}

\begin{figure}[ht]
\centering 
\includegraphics[width=0.9\linewidth]{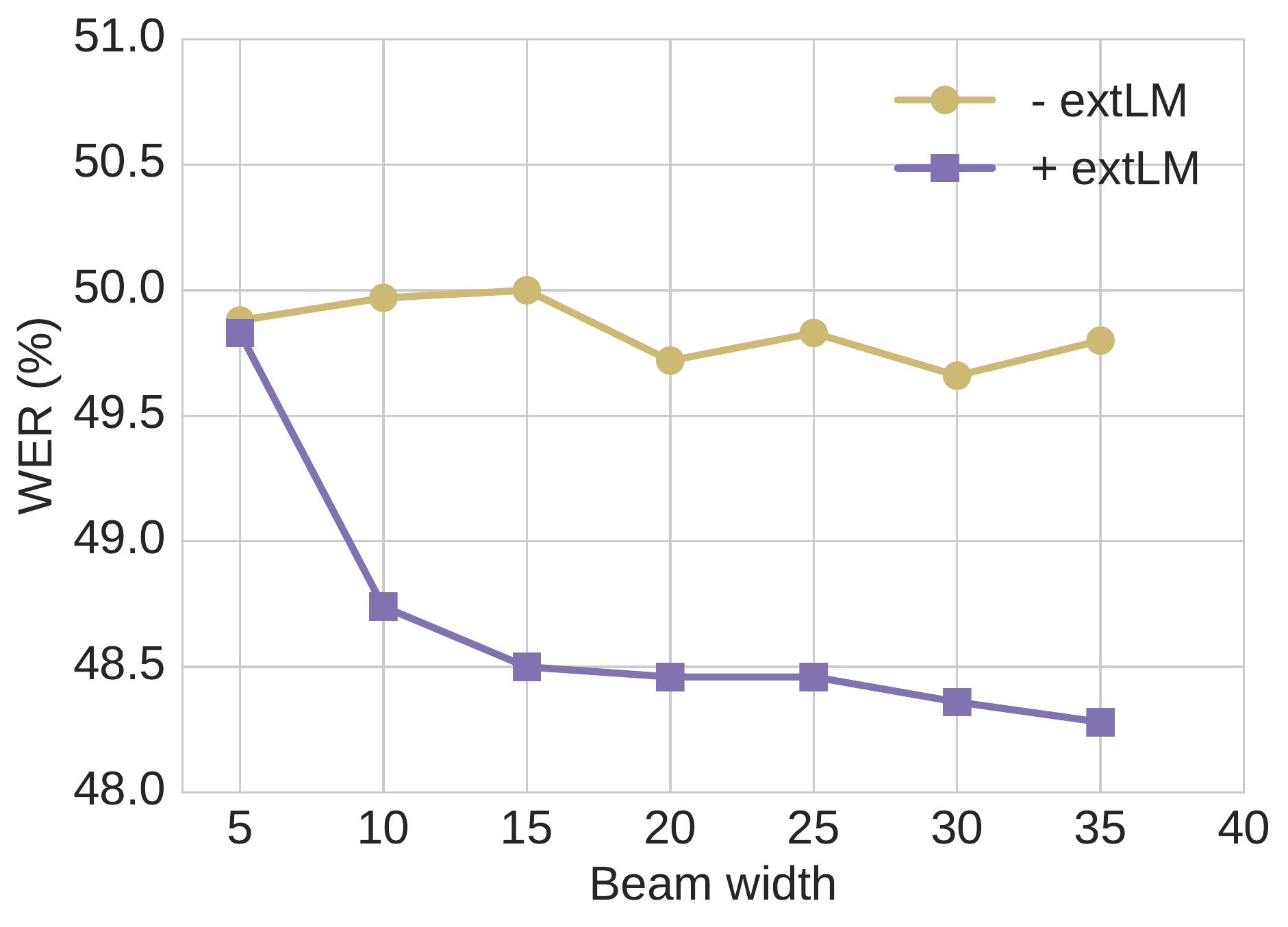}
\caption{The effect of beam width on Word Error Rate for the video-only TM-seq2seq model, when
evaluating on LRS2-BBC.}
\label{fig:beam_plot} 
\end{figure}

In Figure~\ref{fig:wer_per_len} we show how the WER changes as a function of the number of words in a
test sentence.
Figure~\ref{fig:f1_prec_rec} shows the performance of the models on the 30 most common words.
Figure~\ref{fig:beam_plot} shows the effect of increasing
the beam width for the video-only TM-seq2seq model when evaluating on LRS2-BBC. It is noteworthy
that increasing the beam width is more beneficial when decoding with the external language model (+ extLM).

\newpara\noindent\textbf{Decoding examples.}
The model learns to correctly
predict complex unseen
sentences from a wide range of content --
examples are shown in Table~\ref{table:lipres}.

\begin{table}[ht]
\centering
\footnotesize
\begin{tabular}{ l   }
  \toprule
  but this particular reality was not inevitable \\ 
  \midrule
  it would have been completely alien to the rest of london \\ 
  \midrule
  comes from one of the most beautiful parts of the world \\ 
  \midrule
  everyone has gone home happy and that's what it's all about \\ 
  \midrule
  especially when it comes to climate change \\ 
  \midrule
  but it's a different type of animal I want to show you right now \\ 
  \midrule
  but these are one of the most wary birds in the world \\ 
  \midrule
  there's always historical treasures to look at \\ 
  \midrule
  and so how does your brain give you that detail \\ 
  \midrule
  but this is the source of innovation \\ 
  \midrule
  the choices don't make sense because it's the wrong question \\ 
  \midrule
  but it's a global phenomenon \\ 
  \midrule
  mortality is not going down it's going up \\ 
  \bottomrule

\end{tabular} 
\normalsize
\spacetablecapt
\caption{Examples of unseen sentences that TM-seq2seq correctly predicts (video only). 
}
\label{table:lipres}
\end{table}

\subsection{Audio-visual speech recognition}

The visual information can be used to improve the performance of ASR, particularly in environments with background noise~\cite{Mroueh15,Noda15,Petridis18}. Here, we analyse the performance of the audio-visual models described in Section~\ref{sec:arc}.

\newpara\noindent\textbf{Results.}
The results in 
Table~\ref{table:mainresults} demonstrate that
the mouth movements provide important
cues in speech recognition when the audio
signal is noisy; and give an
improvement in performance 
even when the audio signal is clean -- for example the word error rate is
reduced from 10.1\% for audio only to 8.2\%, when
using the audio-visual TM-CTC model. The gains when using the audio-visual TM-seq2seq compared to the audio-only
model are similar.

\newpara\noindent\textbf{Decoding examples.}
Table~\ref{table:avres} shows some of
the many examples where the model fails to
predict the correct sentence from the lips or the audio
alone, but successfully deciphers the words when 
both streams are present.

\begin{table}[ht]
\centering
\footnotesize
\begin{tabular}{ l  l r }
  \toprule
  \textbf{} & \textbf{Transcription} & \textbf{WER \%}  \\ 
  \midrule

 \textbf{GT} & your job needs to be challenging &  \\ 
 \textbf{V} & job is to be challenging & 33 \\ 
 \textbf{A} & your child needs to be challenging & 16 \\
 \textbf{AV} & your job needs to be challenging & 0  \\
  \midrule

 \textbf{GT} & I mean I thought poetry was just self expression &  \\ 
 \textbf{V} & I mean I thought poetry would just suffer as pressure & 44 \\ 
 \textbf{A} & I mean not thought poetry was just self expression & 11 \\ 
 \textbf{AV} & I mean I thought poetry was just self expression & 0 \\ 
  \midrule

 \textbf{GT} & cluster bombs left behind &  \\ 
 \textbf{V} & unless you perhaps have blind & 125 \\ 
 \textbf{A} & close to bombs left behind & 25 \\ 
 \textbf{AV} & cluster bombs left behind & 0 \\ 
  \midrule

 \textbf{GT} & I was the first non family investor in amazon &  \\ 
 \textbf{V} & I was the first not family of us are absurd & 55 \\ 
 \textbf{A} & I was the first non family in bester and amazon & 33 \\ 
 \textbf{AV} & I was the first non family investor in amazon & 0 \\ 
  \bottomrule

\end{tabular} 
\normalsize
\spacetablecapt
\caption{Examples of AVSR results. 
{\bf GT:} Ground Truth;
{\bf A:} Audio only;
{\bf V:} Video only;
{\bf AV:} Audio-visual.
}
\label{table:avres}
\end{table}

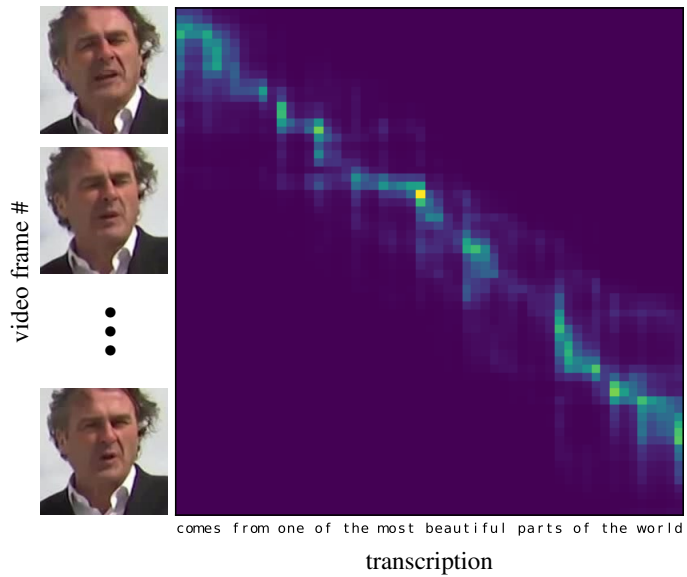
\begin{figure}[t]
\centering 
\begin{subfigure}[b]{0.8\columnwidth}
\begin{tikzpicture}
\hspace{-30pt}
\node (img1)  {
  \includegraphics[width=1\linewidth]{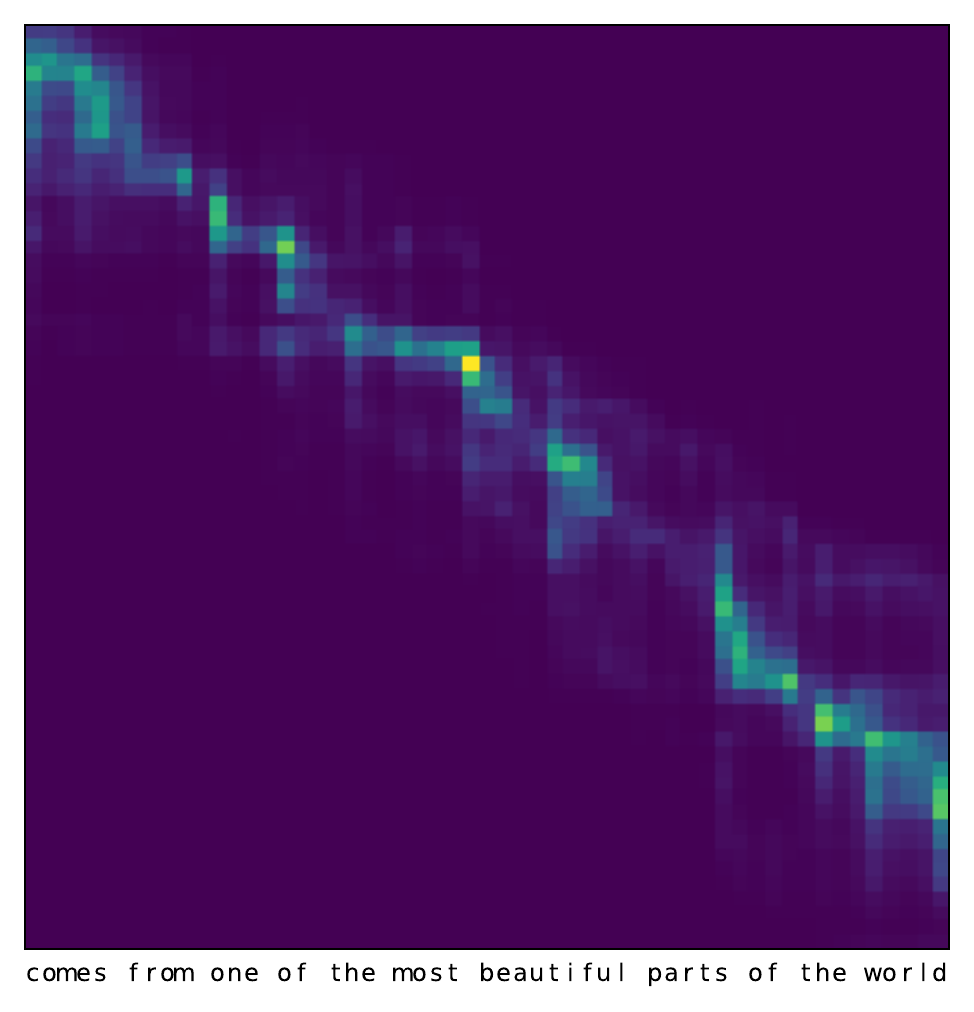}
};
\node [left=of img1, anchor=south,yshift=49pt, xshift=10pt] (img2)  {
  \includegraphics[width=0.237\linewidth]{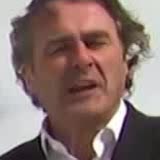}
};
\node [below=of img2,yshift=30pt] (img3)  {
  \includegraphics[width=0.237\linewidth]{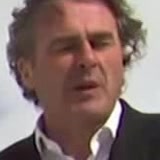}
};
\node [below=of img3, xshift=-3pt,yshift=10pt, rotate=90] (img4)  {
{ $\bullet\bullet\bullet$ }
};
\node [below=of img3,yshift=-8pt] (img5)  {
  \includegraphics[width=0.237\linewidth]{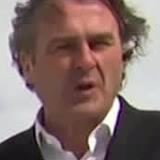}
};
\node[left=of img1, node distance=0cm, rotate=90, anchor=center,yshift=22pt] {video frame \#};
\node[below=of img1, node distance=0cm, yshift=1.2cm] {transcription};
\end{tikzpicture}
\vspace{-10pt}
  \label{fig:align_TM1}
 \end{subfigure}
 \caption{Alignment between the video frames and the character output with TM-seq2seq. The alignment is produced by
 averaging all the encoder-decoder attention heads over all the decoder layers in the log domain.  }

\label{fig:attn} 
\end{figure}

\newpara\noindent{\bf Alignment and attention visualisation.}
The encoder-decoder attention mechanism of the TM-seq2seq model generates explicit alignment
between the input video frames and the hypothesised character output. 
Figure~\ref{fig:attn} visualises the alignment of the 
characters ``comes from one of the most beautiful parts of the world''
and the corresponding video frames. Since the architecture contains multiple attention heads, we
obtain the alignment by averaging the attention masks over all the decoder layers in the log domain.

\newpara\noindent\textbf{Noisy audio.}
We perform the audio-only and audio-visual experiments with noisy audio, synthesized by adding babble noise to the original utterances.
Speech recognition in a noisy environment is extremely challenging,
as can be seen from the significantly lower performance of the off-the-shelf Google S2T ASR baseline (over 60\% performance degradation compared to clean). This difficulty is
also reflected on the performance of our audio-only models, that the word error rates
similar to the ones obtained when only using the lips.
However combining the two modalities provides a significant improvement, with the
word error rate dropping significantly, by up to $30\%$. Notably, the audio-visual models
perform much better than either the video-only, or audio-only ones under the presence of loud background
noise. 

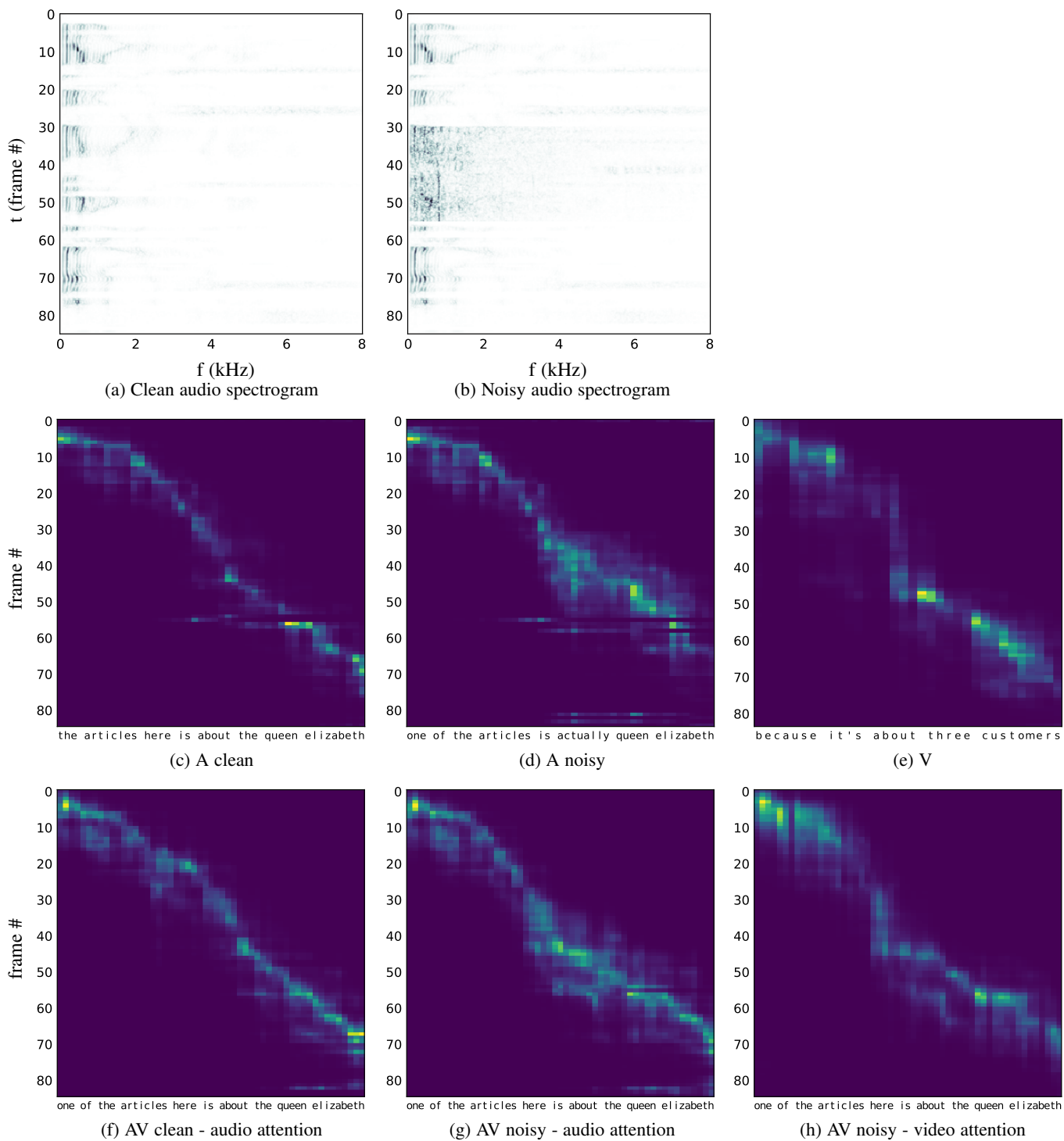
\begin{figure*}[t]

\begin{subfigure}[b]{0.66\columnwidth}
\begin{tikzpicture}
\node (img1)  {
  \includegraphics[width=1\columnwidth]{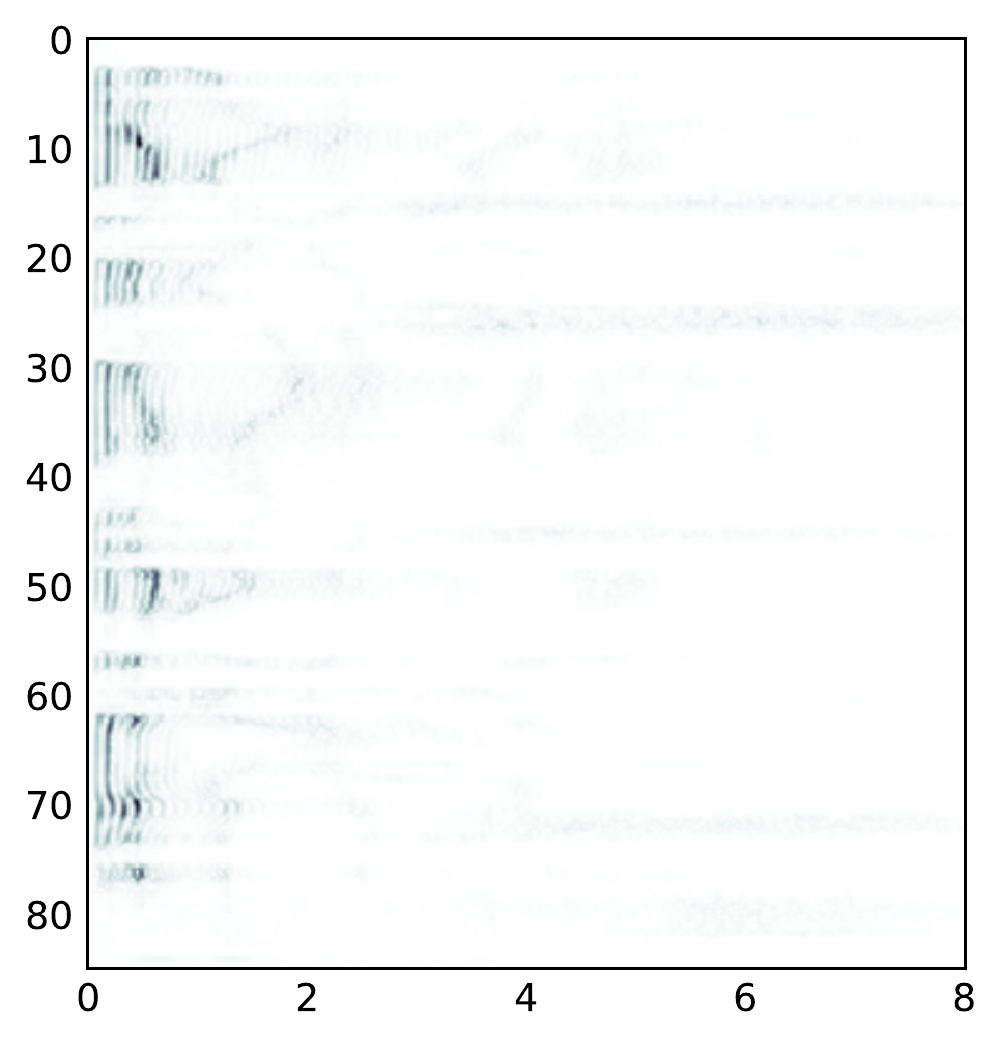}
};
\node[left=of img1, node distance=0cm, rotate=90, anchor=center,yshift=-0.9cm] {t (frame \#)};
  \node[below=of img1, node distance=0cm, yshift=1.2cm, xshift=0.4cm] {f (kHz)};
\end{tikzpicture}
\vspace{-20pt}
  \caption{Clean audio spectrogram\hspace*{-4em}}
  \label{fig:align_TM2}
 \end{subfigure}
\begin{subfigure}[b]{0.66\columnwidth}
\begin{tikzpicture}
\node (img1)  {
  \includegraphics[width=1\linewidth]{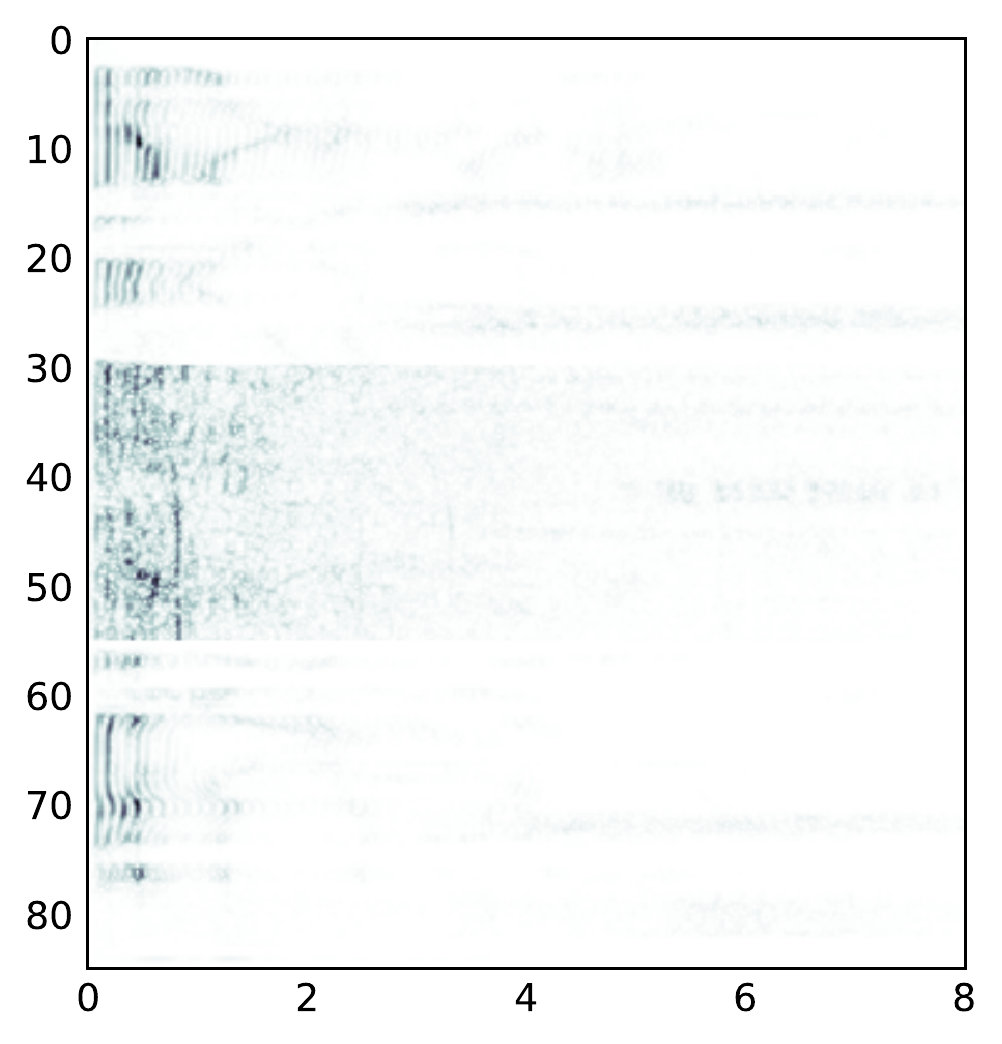}
};
\node[] () [left of = img1, xshift=-2.25cm] {};
\node[below=of img1, node distance=0cm, yshift=1.2cm, xshift=0.4em] {f (kHz)};
\end{tikzpicture}
\vspace{-20pt}
  \caption{Noisy audio spectrogram\hspace*{-4em}}
  \label{fig:align_TM3}
 \end{subfigure}
 \hspace{5.9cm} 

\begin{subfigure}[b]{0.66\columnwidth}
\begin{tikzpicture}
\node[left=of img1, node distance=0cm, rotate=90, anchor=center,yshift=-0.9cm] { frame \# };
\node (img1)  {
  \includegraphics[width=1\columnwidth]{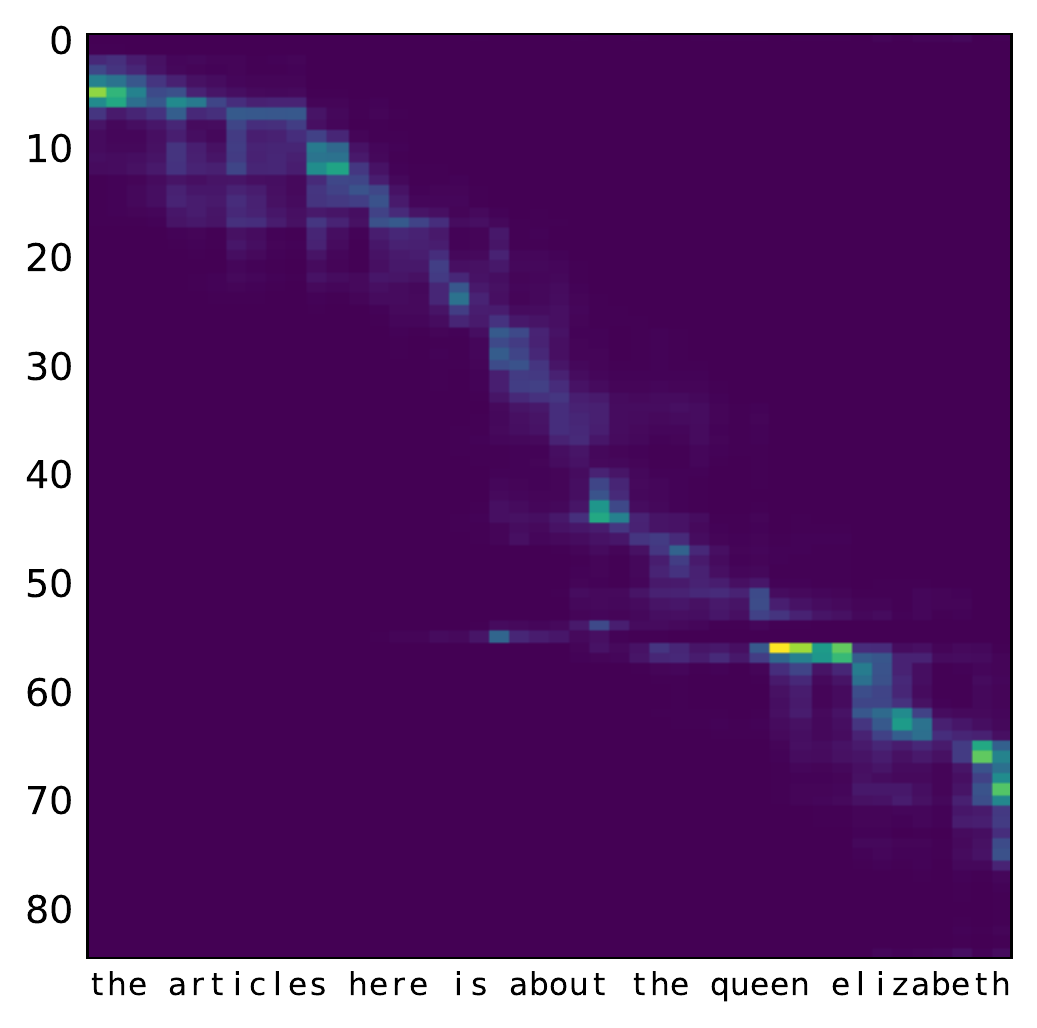}
};
\end{tikzpicture}
\vspace{-20pt}
  \caption{A clean\hspace*{-4em}}
  \label{fig:align_TM4}
 \end{subfigure}
\begin{subfigure}[b]{0.66\columnwidth}
\begin{tikzpicture}
\node (img1)  {
  \includegraphics[width=1\linewidth]{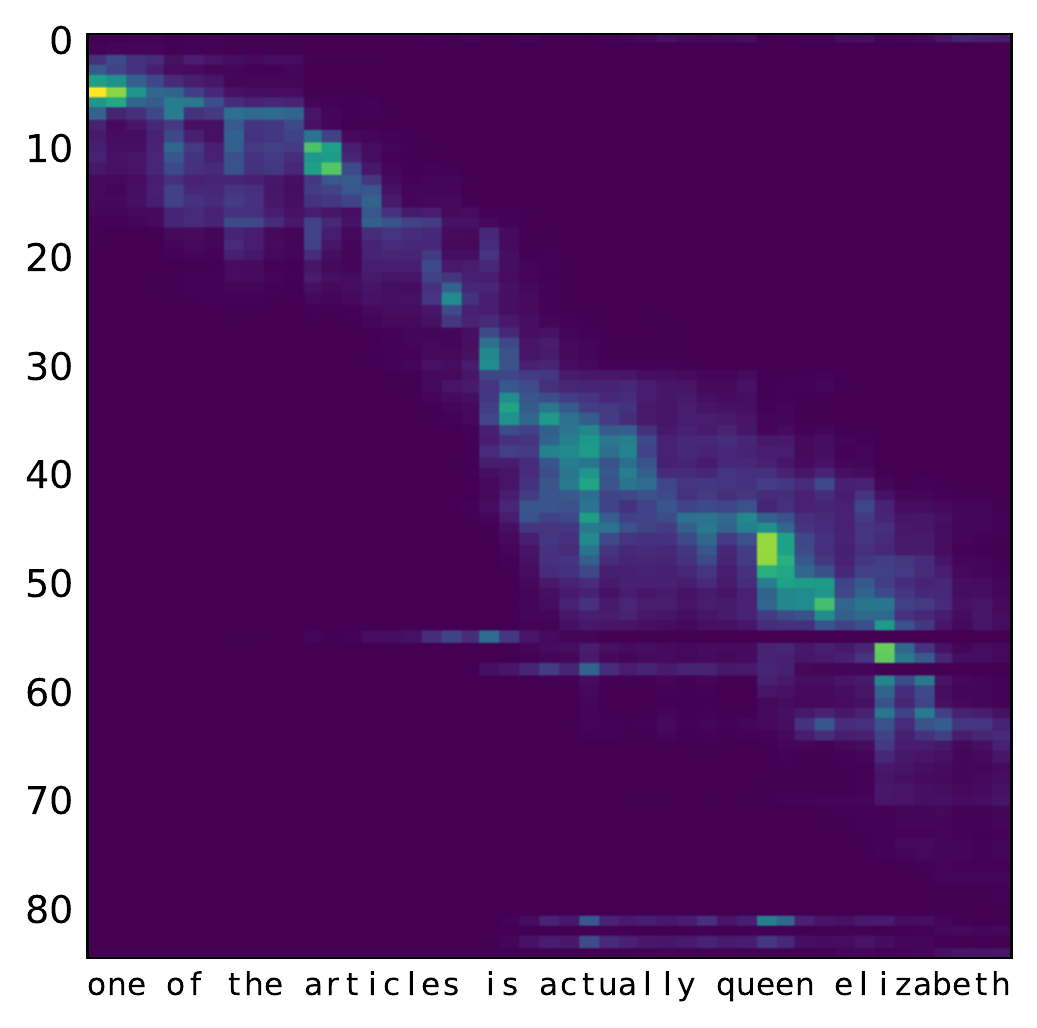}
};
\node[] () [left of = img1, xshift=-2.25cm] {};
\end{tikzpicture}
\vspace{-20pt}
  \caption{A noisy\hspace*{-4em}}
  \label{fig:align_TM5}
 \end{subfigure}
\begin{subfigure}[b]{0.66\columnwidth}
\begin{tikzpicture}
\node (img1)  {
  \includegraphics[width=1\linewidth]{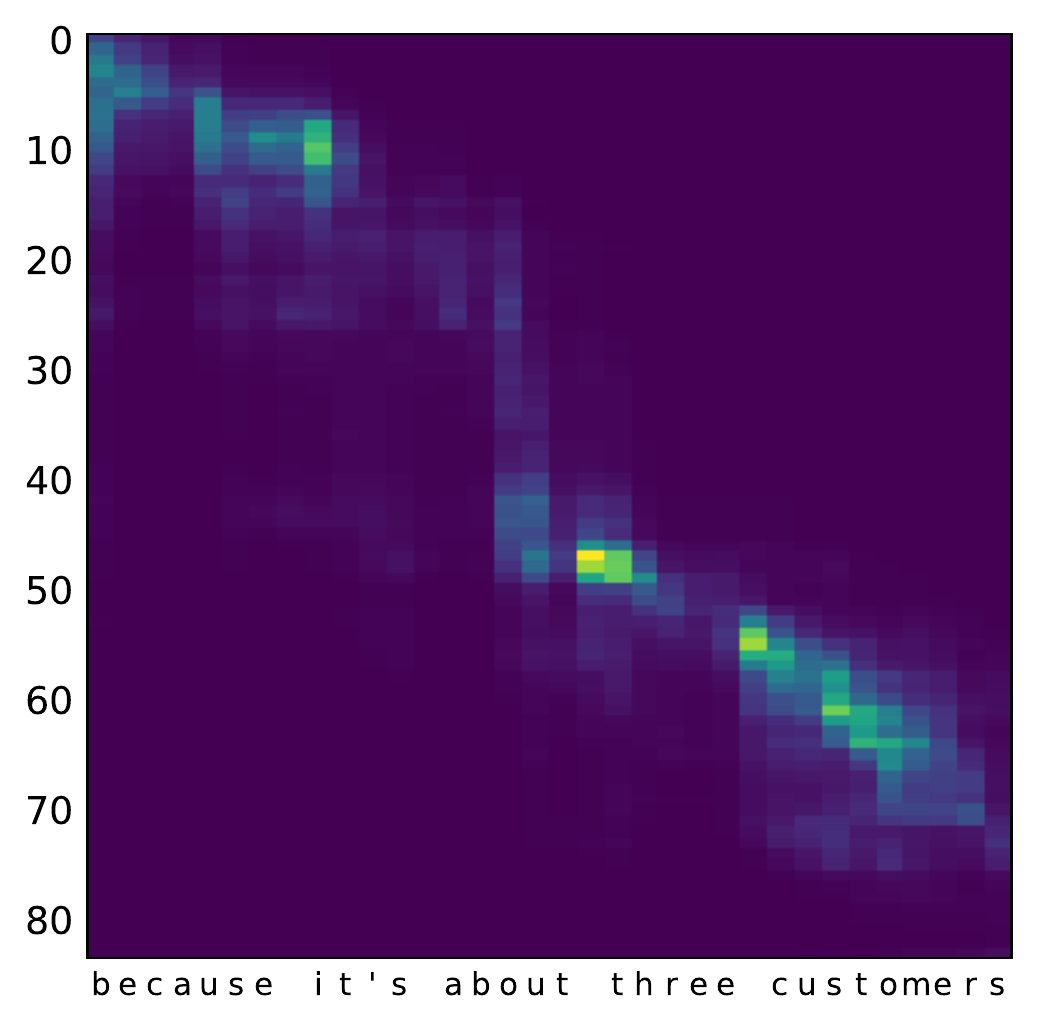}
};
\node[] () [left of = img1, xshift=-2.2cm] {};
\end{tikzpicture}
\vspace{-20pt}
  \caption{V\hspace*{-4em}}
  \label{fig:align_TM6}
 \end{subfigure}

\begin{subfigure}[b]{0.66\columnwidth}
\begin{tikzpicture}
\node[left=of img1, node distance=0cm, rotate=90, anchor=center,yshift=-0.9cm] {frame \# };
\node (img1)  {
  \includegraphics[width=1\linewidth]{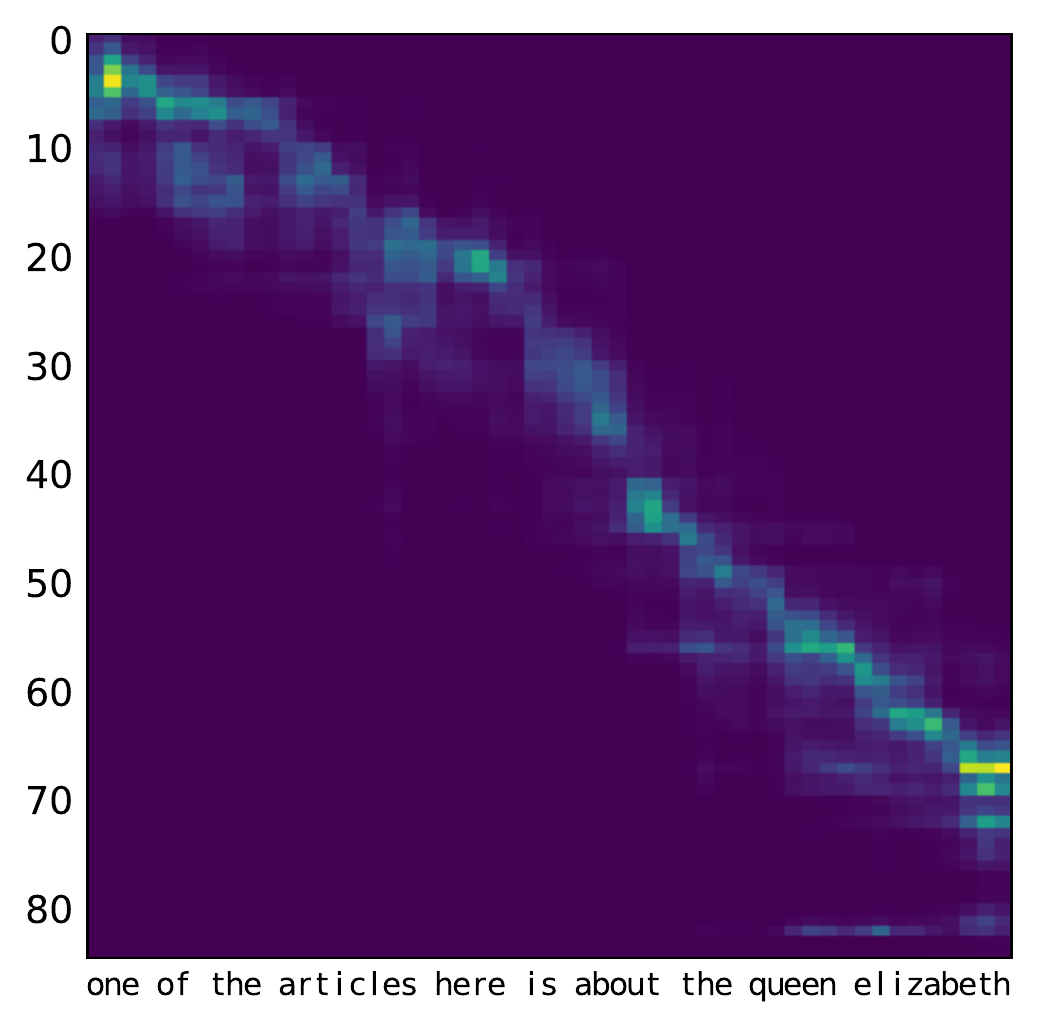}
};
\end{tikzpicture}
\vspace{-20pt}
  \caption{AV clean - audio attention\hspace*{-4em}}
  \label{fig:align_TM7}
 \end{subfigure}
\begin{subfigure}[b]{0.66\columnwidth}
\begin{tikzpicture}
\node (img1)  {
  \includegraphics[width=1\linewidth]{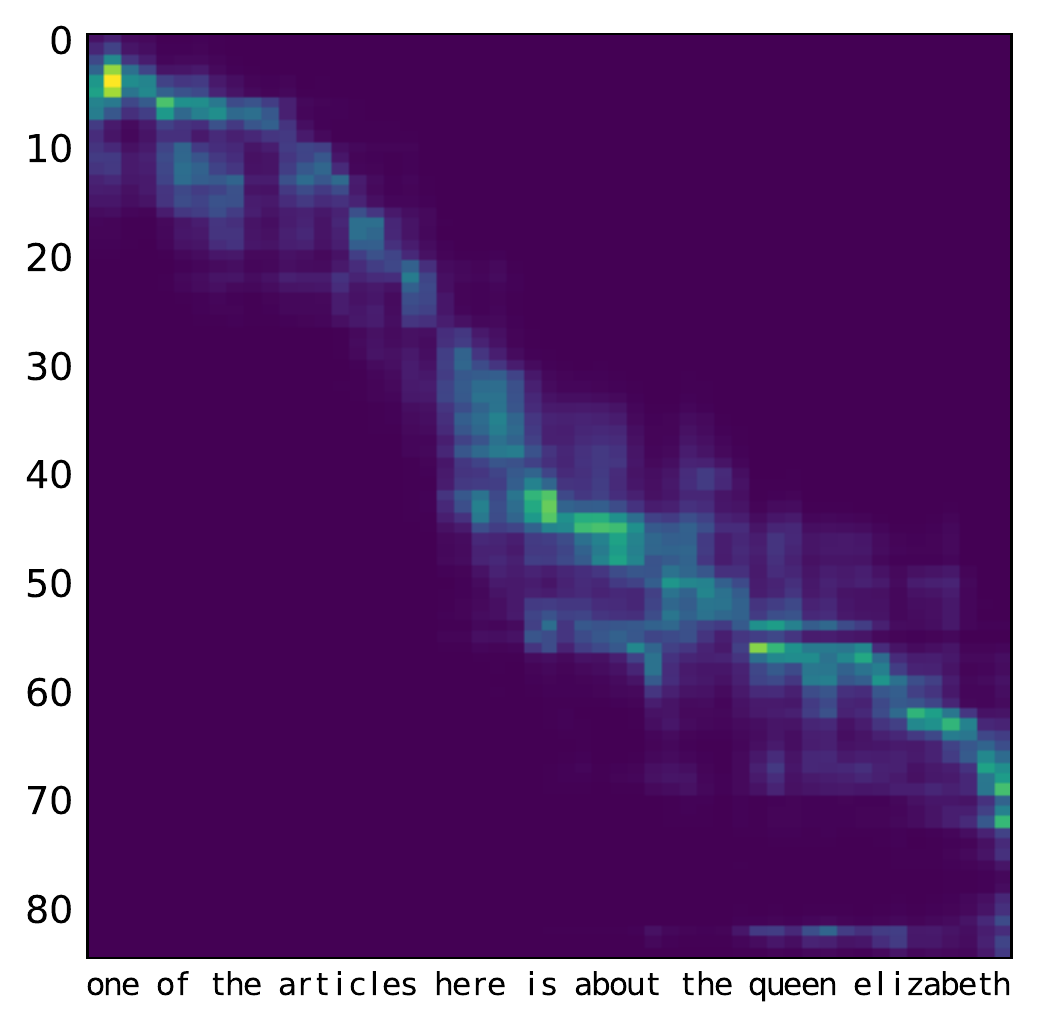}
};
\node[] () [left of = img1, xshift=-2.25cm] {};
\end{tikzpicture}
\vspace{-20pt}
  \caption{AV noisy - audio attention\hspace*{-4em}}
  \label{fig:align_TM8}
 \end{subfigure}
\begin{subfigure}[b]{0.66\columnwidth}
\begin{tikzpicture}
\node (img1)  {
  \includegraphics[width=1\linewidth]{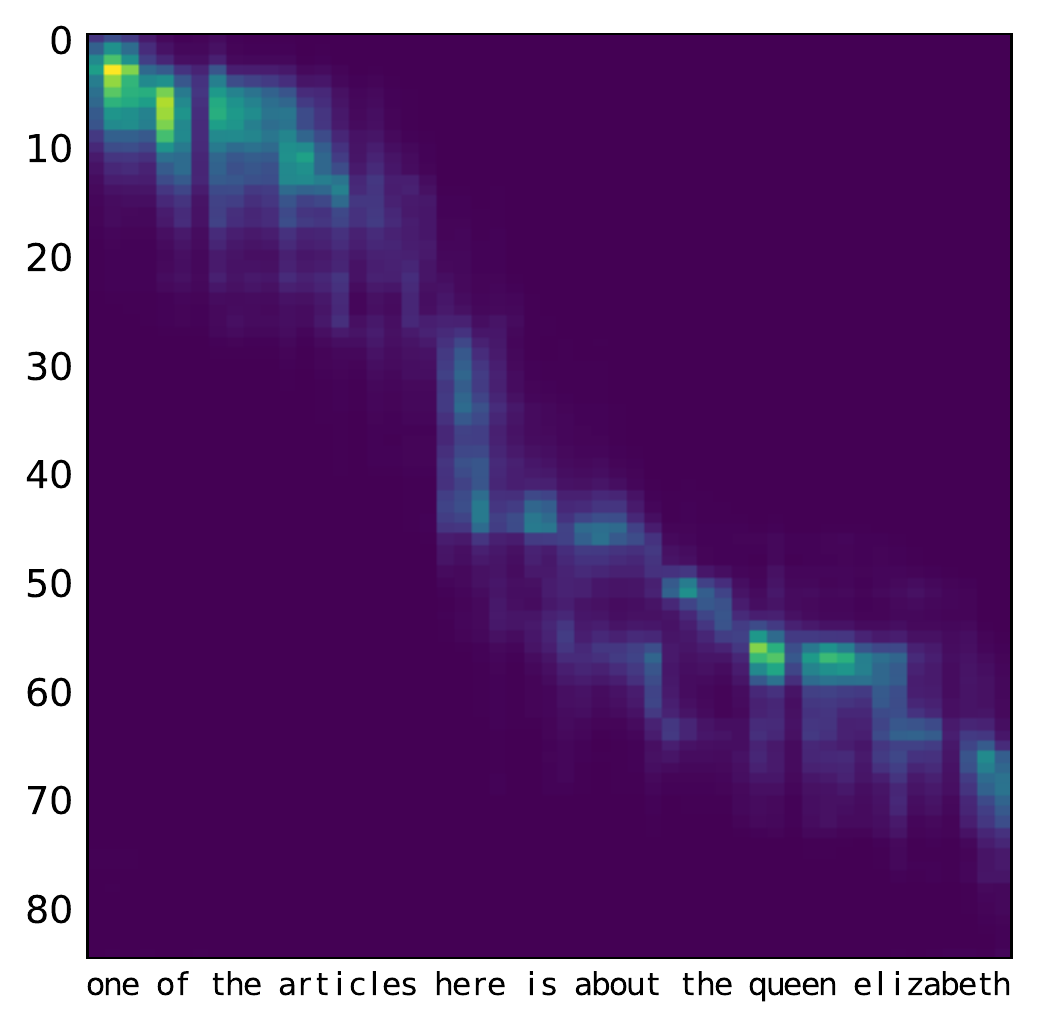}
};
\node[] () [left of = img1, xshift=-2.2cm] {};
\end{tikzpicture}
\vspace{-20pt}
  \caption{AV noisy - video attention\hspace*{-4em}}
  \label{fig:align_TM9}
 \end{subfigure}

 \caption{
   Visualization of the effect of additive noise on the attention masks of the different TM-seq2seq models. We show the attentions
   on (a) the clean audio utterance, and (b) on the noisy utterance which we obtain by adding babble
   noise to the 25 central audio frames. 
   Comparing (c) with (d), the attention of the audio-only models appears to be more spread around the area where the noise
   is applied, while the last frames are not attended upon. Similarly for the audio-visual model, the audio attention
   is more focused when the audio is clean (f) compared to when it is noisy (g).
   The ground truth transcription of the sentence is ``one of the articles there is about the queen elizabeth''.
   Observing the transcriptions, we see that the audio-only model (d) does not predict the central words correctly when noise is added, however
   the audio-visual model (g \& h) successfully transcribes the sentence, by leveraging the visual cues.
   Interestingly, in this particular example, the transcription that the video-only model outputs
   (e) is completely wrong; the combination of both modalities however yields a correct prediction.  
   Finally, the attention mask of the AV model on the video input (f) has a clear monotonic
   trend and is similar to the one of the video-only model (e); this also
   verifies that the model indeed learns to use the video modality even when audio is present.  }

\label{fig:attn_compare} 
\end{figure*}

\newpara\noindent\textbf{AV attention visualization.}
In Figure~\ref{fig:attn_compare} we compare the attention masks of different TM-seq2seq models in the
presence and absence of additive babble noise in the audio stream.


\subsection{Out-of-sync audio and video}
\begin{figure}[ht]
\centering 
\includegraphics[width=1\linewidth]{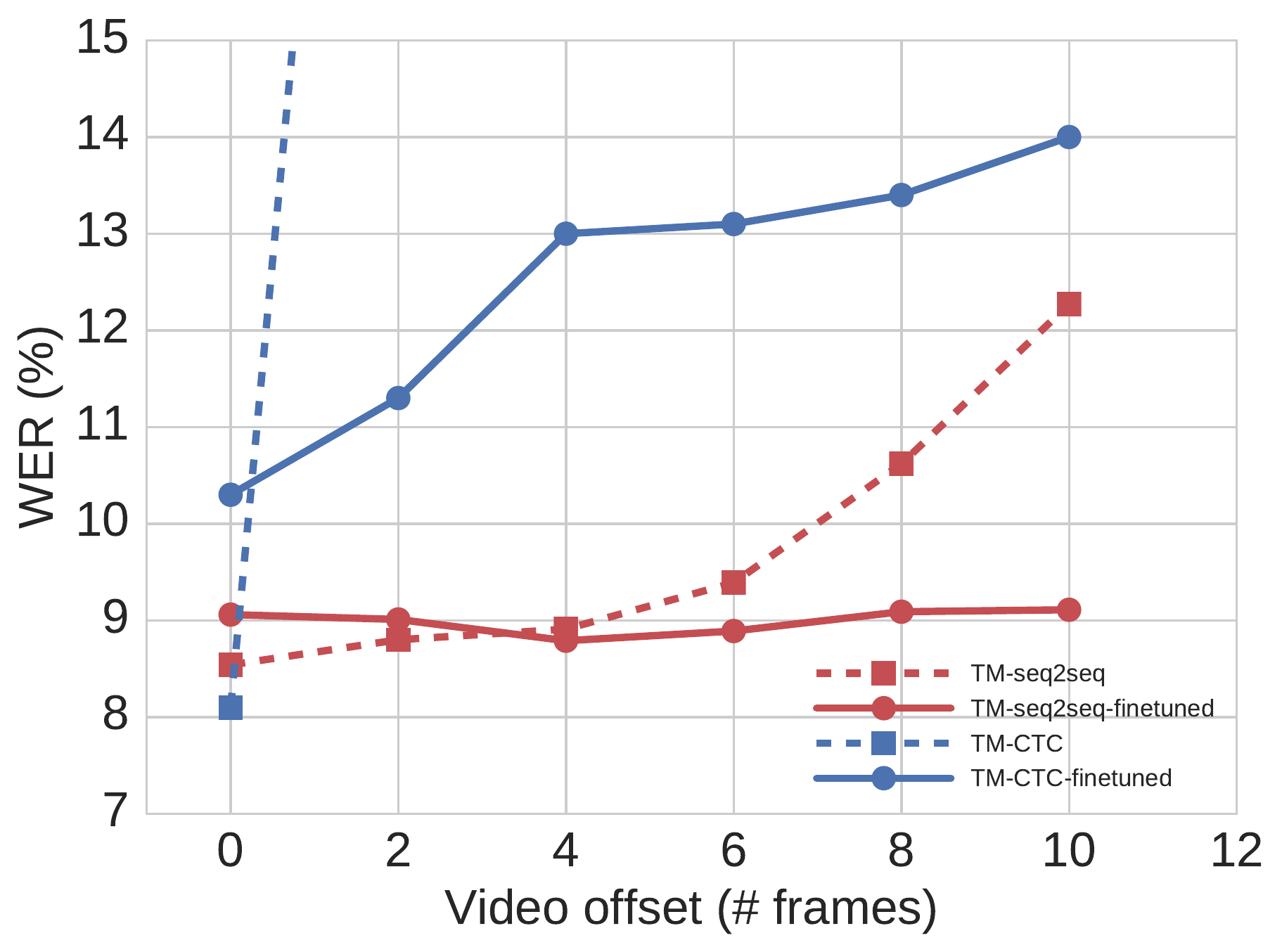}
 \caption{WER scored by the audio-visual models on LRS2-BBC when the video frames are artificially shifted by 
 a number of frames compared to audio. The TM-seq2seq model is only fine-tuned for one epoch, while
 CTC for 4 epochs on the train-val set. }
\label{fig:sync} 
\end{figure}

Here, we assess the performance of the audio-visual models when the audio and video inputs are not temporally aligned.
Since the audio and video have been synchronised in our dataset, we synthetically shift
the video frames to achieve an out-of-sync effect. 
We evaluate the performance on de-synchronised samples of the LRS2-BBC dataset.
We consider the TM-CTC and TM-seq2seq architectures, with and without fine-tuning on randomly
shifted samples.
The results are shown in Figure \ref{fig:sync}. 
It is clear that the TM-seq2seq architecture is more resistant to these shifts. We only need to
calibrate the model for one epoch for the out-of-sync effect to practically vanish. This showcases the
advantage of employing independent encoder-decoder attention mechanisms for the two modalities.
In contrast, TM-CTC, that concatenates the two encodings, struggles to deal with the shifts, even
after several epochs of fine-tuning.

\subsection{Discussion on seq2seq vs CTC}

The TM-seq2seq model performs significantly better for lip-reading in terms of WER, when no audio is supplied. For audio-only
or audio-visual tasks, the two methods perform similarly. However the CTC models
appear to handle background noise better; in the presence of loud babble noise, both the audio-only
and audio-visual TM-seq2seq models perform
significantly worse that their TM-CTC counterparts.

\newpara\noindent\textbf{Training time}.
The TM-seq2seq models have a more complex architecture and are harder to train, with the full
audio-visual model taking approximately 8 days to complete the full curriculum for both datasets,
on a single GeForce Titan X GPU with 12GB memory.
In contrast, the audiovisual TM-CTC model trains faster i.e. in approximately 5 days on the same
hardware.
It should be noted however that since both architectures contain no recurrent modules and no batch
normalization, their implementation can be heavily parallelized into multiple GPUs.

\newpara\noindent\textbf{Inference time}.
Decoding of the TM-CTC model does not require auto-regression and therefore the CTC
probabilities need only be evaluated once, regardless of the beam width $W$.
This is not the case for TM-seq2seq, where for every step of the beam search, the decoder subnetwork needs
to be evaluated $W$ times. 
This makes the decoding of the CTC model faster, which can be an important factor for deployment.

\newpara\noindent\textbf{Language modelling}.
Both models perform better when an external language model is incorporated in the beam search,
however the gains are much higher for TM-CTC, since no explicit
language consistency is enforced by the visual model alone.

\newpara\noindent\textbf{Generalization to longer sequences}.
We observed that the TM-CTC model generalizes better and adapts faster as the sequence lengths are
increased during the curriculum learning. We
believe this also affects the training time as the latter takes more epochs to converge.

\section{Conclusion}
\label{sec:conc}
\spacesection

In this paper, we introduced a large-scale, unconstrained audio-visual dataset, LRS2-BBC,
formed by collecting and preprocessing thousands of videos from the British television.

We considered two models that can transcribe 
audio and video sequences of speech into characters and showed that the same architectures can also be used when only one
of the modalities is present.
Our best visual-only model surpasses the
performance of the previous state-of-the-art  on the LRS2-BBC lip reading dataset
by a large margin, and sets a strong baseline for the recently released LRS3-TED. We finally demonstrate that visual
information helps improve speech recognition performance even when
the clean audio signal is available.  Especially in the presence of noise in the audio, combining the two modalities
leads to a significant improvement. 


\clearpage

\bibliographystyle{ieee}
\bibliography{longstrings,vgg_local,vgg_other,mybib}



\clearpage


%

\appendices
\section{Visual front-end architecture}  \label{ap:visual_frontend}

The details of the spatio-temporal front-end are given in Table~\ref{tab:resnet_arch}.

\begin{table}[h]
\begin{center}
\scriptsize
\begin{tabular}{ lll } 
 \toprule
 \textbf{Layer Type} & \textbf{Filters} & \textbf{Output dimensions} \\ 
 \midrule
 Conv 3D & $5\times7\times7$, $64$, $/ [1, 2, 2]$  & $T\times \frac{H}{2} \times \frac{W}{2} \times 64$ \\
\addlinespace[0.5em]
 Max Pool 3D & $/ [1, 2, 2]$                       & $T\times \frac{H}{4} \times \frac{W}{4} \times 64$ \\
 \midrule
 Residual Conv 2D  & [$3\times3$, $64$] $\times2$  $/  1$  & $T\times \frac{H}{4} \times \frac{W}{4} \times 64$ \\
\addlinespace[0.5em]
 Residual Conv 2D  & [$3\times3$, $64$] $\times2$  $/  1$  & $T\times \frac{H}{4} \times \frac{W}{4} \times 64$ \\
 \midrule                                                              
 Residual Conv 2D  & [$3\times3$, $128$] $\times2$  $/ 2$  & $T\times \frac{H}{8} \times \frac{W}{8} \times 128$ \\
\addlinespace[0.5em]
 Residual Conv 2D  & [$3\times3$, $128$] $\times2$  $/ 1$  & $T\times \frac{H}{8} \times \frac{W}{8} \times 128$ \\
 \midrule                                                             
 Residual Conv 2D  & [$3\times3$, $256$] $\times2$  $/ 2$  & $T\times \frac{H}{16} \times \frac{W}{16} \times 256$ \\
\addlinespace[0.5em]
 Residual Conv 2D  & [$3\times3$, $256$] $\times2$  $/ 1$  & $T\times \frac{H}{16} \times \frac{W}{16} \times 256$ \\
 \midrule                                                             
 Residual Conv 2D  & [$3\times3$, $512$] $\times2$  $/ 2$  & $T\times \frac{H}{32} \times \frac{W}{32} \times 512$ \\
\addlinespace[0.5em]
 Residual Conv 2D  & [$3\times3$, $512$] $\times2$  $/ 1$  & $T\times \frac{H}{32} \times \frac{W}{32} \times 512$ \\
 \midrule  

\end{tabular}
\normalsize
\end{center}
\caption{Architecture details for the spatio-temporal visual front-end \cite{Stafylakis17}. The
strides for the residual 2D convolutional blocks apply to the first layer of the block only (i.e. the
total down-sampling factor in the network is 32). A short cut connection is added after every pair of
2D convolutions \cite{He15}. The 2D convolutions are applied separately on every time-frame.}
\label{tab:resnet_arch}
\end{table}


\section{Transformer architecture details} \label{ap:transformer_details}

\begin{figure}[] 
        \vspace{8pt}
    \centering
            \includegraphics[width=\linewidth]{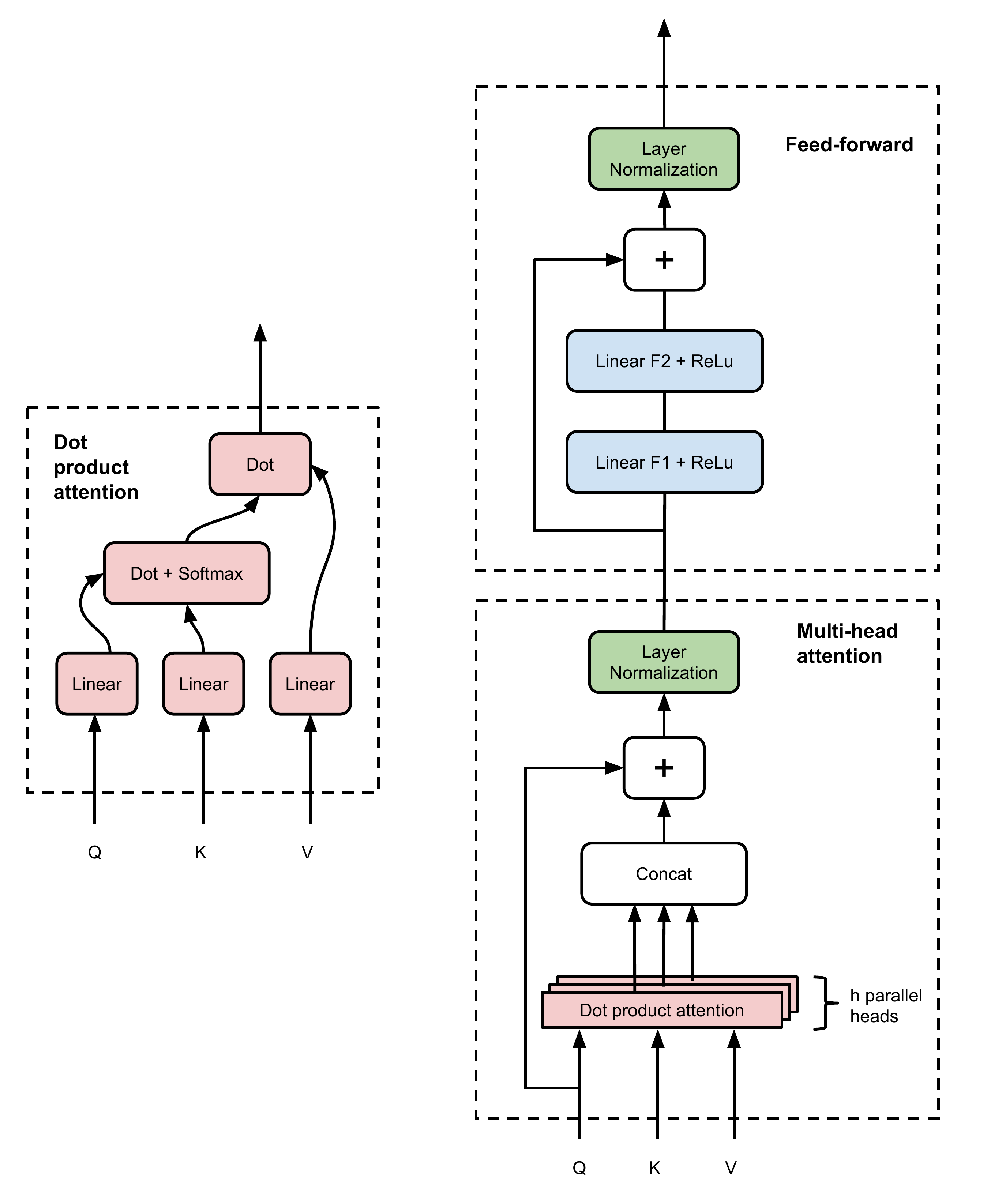}
            \caption{ Details of multi-head attention building blocks }
            \label{fig:transformer_small}
\end{figure}

The details of the building blocks used by our models are outlined in Figure~\ref{fig:transformer_small}.
The same multi-head attention block shown is used for both the self-attention and encoder-decoder
attention layers of the models.
A multi-head attention block, as described by Vaswani {\it et al.}~\cite{Vaswani2017},
receives a query ($Q$), a key ($K$) and a value ($V$) tensor as inputs and produces
$h$ context vectors, one for every attention head $i$:
\begin{equation}
  Att_i(Q,K,V) =  softmax( \frac{ (W_i^q Q^T)^T (W_i^k K^T) } { \sqrt{d_k} } ) (W_i^v V^T)^T   \nonumber
\end{equation}
where $Q$, $K$, and $V$ have size $d_{model}$ and $dk$ =  $\frac{d_{model}}{h}$ is the size of every attention head.
The $h$ context vectors are concatenated and propagated through a feedforward block that consists of
two linear layers with ReLU non-linearities in between.
For the self-attention layers it is always $Q=K=V$, while for the encoder-decoder attention of the
TM-seq2seq model,
$K$ = $V$ are the encoder outputs to be attended upon and $Q$ is the decoder input, i.e. the
network's output at the previous decoding step for the first layer and the output of the previous decoder layer for the rest.
We use the same architecture hyperparameters as the original base model of Vaswani {\it et al.}~\cite{Vaswani2017}
with $d_{model} = 512$ and $h=8$ attention heads
everywhere. The sizes of the two linear layers in the feedforward block are $F1=2048$,  $F2=512$.

\section{Seq2Seq decoding with external language model} \label{ap:seq2seq_dec}
For decoding with the TM-seq2seq model, we use a left-to right beam search with width $W$ as in~\cite{Kannan17,Wu16},
with the hypotheses $y$ being scored as follows: 
\begin{equation}
  score(x,y) = \frac { log \ p(y|x) + \alpha \ log \ p_{LM}(y) }{ LP(y) } \nonumber
\end{equation}
where $p(y|x)$ and $p_{LM}(y)$ are the probabilities obtained from the visual and language models respectively and
LP is a length normalization factor LP(y) = $ \Big( \frac{5+|y|}{6}\Big)^\beta $ \cite{Wu16}. We did not experiment
with a coverage penalty.
The best values for the hyperparameters were determined via grid search on the validation set: for
decoding without the external language model they were set to  $W=6$, $\alpha=0.0$,
$\beta=0.6$ and for decoding with the external language model (+ extLM) to $W=35$, $\alpha=0.1$ $\beta=0.7$.

\section{CTC decoding algorithm with external language model} \label{ap:ctc_dec}

Algorithm~\ref{alg:beam_lm} describes the CTC decoding procedure with an external language model.
It is also a beam search with width W and hyperparameters $\alpha$ and $\beta$ that control the
relative weight given to the LM and the length penalty. The beam search is similar to the one
described for seq2seq above, with some additional bookkeeping required to handle the emission of
repeated and blank characters and normalization LP(y) = $ |y|^\beta$.
We obtain the best results on the validation set with $W=100$, $\alpha=0.5$, $\beta=0.1$.

\begin{algorithm}[]
  \caption{CTC Beam search decoding with Language Model adapted from \cite{Maas15}. 
    Notation: 
    A is the alphabet;
    $p_b(s,t)$ and $p_{nb}(s,t)$ are the probabilities of partial output transcription s resulting from paths
  ending in blank and non-blank token respectively, given the input sequence up to time $t$; $p(s,t) = p_b(s,t) + p_{nb}(s,t)$. }
   \label{alg:beam_lm}
   \hskip -2em
\begin{algorithmic} 

  \State \textbf{Parameters} CTC probabilities $p^{ctc}_{1:T}$, word dictionary, beam width $W$,
  hyperparameters $\alpha$, $\beta$

  \State initialize $\mathbf{B_{t}} \leftarrow$ \{$\varnothing$\}; 
  $\mathbf{p_b(\varnothing, 0)} \leftarrow$ 1; $\mathbf{p_{nb}(\varnothing, 0)} \leftarrow$ 0 

  \For{$t=1$ {\bfseries to} $T$} 

    \State $\mathbf{B_{t-1}} \leftarrow$ $W$ prefixes with highest $\frac{ log \ p(s,t)}{|s|^\beta}$ in $\mathbf{B}_t$
    \State $\mathbf{B_t} \leftarrow$ \{\}

    \For{prefix $s$ {\bfseries in} $\mathbf{B_{t-1}}$} 

      \State $ c^{-} \leftarrow  $ last character of $s$

      \State  $p_b(s,t) \leftarrow p^{ctc}_t(-,t) p(s,t-1) $  \Comment{adding a blank}
      \State $p_{nb}(s,t) \leftarrow p^{ctc}_t(c^-,t) p_{nb}(s,t-1)$  \Comment{repeated}

      \State add $s$ to $\mathbf{B_{t}}$

      \For{character $c$ {\bfseries in} $A$} 

        \State $ s^{+} \leftarrow  s + c$
        
        \If {$s$ does not end in $c$} 
          \State  $p_{c} \leftarrow p^{ctc}_t(c,t) p(s,t-1) p_{LM}(c|s)^\alpha $  
        \Else
          \State \Comment{repeated chars must have blanks in between}
          \State  $p_{c} \leftarrow p^{ctc}_t(c,t) p_{b}(s,t-1) p_{LM}(c|s)^\alpha $  
        \EndIf 

        \If {$s^{+}$ is already in $\mathbf{B_t}$ }
        \State $p_{nb}(s^+,t) \leftarrow p_{nb}(s^+,t) + p_{c}$

        \Else
          \State add $s^+$ to $\mathbf{B_t}$
          \State $p_{nb}(s,t) \leftarrow 0$ 
          \State $p_{nb}(s^+,t) \leftarrow p_{c}$
        \EndIf

      \EndFor

    \EndFor

\EndFor
        
\State \Return $ max_{s \in B_t} \frac{ log \ p(s,T)}{|s|^\beta}$ in $\mathbf{B}_T$

\end{algorithmic}
\end{algorithm}

\section{Precision and Recall from edit distance} \label{ap:prec_recall}
The F1, precision and recall rates shown in figure \ref{ap:prec_recall}, are calculated from the
word-wise minimum edit distance operations.
For every sample in the evaluation set we can calculate the fewest word substitution,
insertion and deletion operations needed to get from the ground truth to the predicted
transcription. After aggregating those operations over
the evaluation set for every word, we calculate the average measures per word as follows:
\begin{align}
  TP(w) = n_{m}(w) \nonumber \\
  FN(w) =  \sum_j n_{s}(j,w) + n_{i}(w) \nonumber \\ 
  FP(w) =  \sum_j n_{s}(w,j)+ n_{d}(w) \nonumber \\
  Precision(w) = \frac{TP(w)}{TP(w)+FP(w)} \nonumber \\
  Recall(w) = \frac{TP(w)}{TP(w)+FN(w)} \nonumber \\
  F1(w) = 2\frac{Precision(w) Recall(w)}{Precision(w) + Recall(w)}  \nonumber 
\end{align}
where $n_{s}$(w,j) is the total count over the evaluation set of substitutions of word $j$ with word $w$,
and $n_{m}$(w), $n_{i}$(w) and $n_{d}$(w) are the total matches, deletions and insertions respectively of word
$w$.

\ifCLASSOPTIONcompsoc
  \section*{Acknowledgments}
\else
  \section*{Acknowledgment}
\fi

Funding for this research is provided by the EPSRC 
Programme Grant Seebibyte EP/M013774/1, the EPSRC
CDT in Autonomous Intelligent Machines and Systems, 
and the Oxford-Google DeepMind Graduate Scholarship. We are very grateful to Rob Cooper and Matt 
Haynes at BBC Research for help in obtaining the dataset. We would like to thank Ankush Gupta
for helpful comments and discussion.

\ifCLASSOPTIONcaptionsoff
  \newpage
\fi

\end{document}